\documentclass{article} % For LaTeX2e
\usepackage[preprint]{colm2026_conference}

\usepackage{microtype}
\usepackage{hyperref}
\usepackage{url}
\usepackage{booktabs}
\usepackage{tabularx}
\usepackage{graphicx}
\usepackage{amsmath}
\usepackage{tcolorbox}
\usepackage{algorithm}
\usepackage{algpseudocode}
\usepackage{enumitem}
\usepackage{wrapfig}
\usepackage{caption} % Required for \captionof
\usepackage{multirow}
\usepackage{float}
\usepackage{amssymb}% http://ctan.org/pkg/amssymb
\usepackage{pifont}% http://ctan.org/pkg/pifont
\usepackage{siunitx}
\usepackage{mdframed}
\usepackage{parskip}
\usepackage{courier}
\usepackage{multirow}
\usepackage{subcaption}
\usepackage{makecell}
\usepackage[T1]{fontenc}

\newcommand{\cellacc}[2]{\begin{tabular}[c]{@{}c@{}}#1\\[-1pt]\scriptsize(#2)\end{tabular}}

\algrenewcommand\algorithmiccomment[1]{\hfill\textcolor{gray}{\small\textit{// #1}}}

\definecolor{sysblue}{RGB}{220,232,248}
\definecolor{dyngreen}{RGB}{220,245,220}
\definecolor{adaptamber}{RGB}{255,243,210}
\definecolor{retryred}{RGB}{252,228,224}
\definecolor{borderblue}{RGB}{70,130,180}
\definecolor{bordergreen}{RGB}{60,150,60}
\definecolor{borderamber}{RGB}{200,140,0}
\definecolor{borderred}{RGB}{180,60,50}

\newmdenv[
  backgroundcolor=sysblue,
  linecolor=borderblue,
  linewidth=1.2pt,
  innerleftmargin=10pt, innerrightmargin=10pt,
  innertopmargin=6pt, innerbottommargin=6pt
]{sysbox}

\newmdenv[
  backgroundcolor=dyngreen,
  linecolor=bordergreen,
  linewidth=1.2pt,
  innerleftmargin=10pt, innerrightmargin=10pt,
  innertopmargin=6pt, innerbottommargin=6pt
]{dynbox}

\newmdenv[
  backgroundcolor=adaptamber,
  linecolor=borderamber,
  linewidth=1.2pt,
  innerleftmargin=10pt, innerrightmargin=10pt,
  innertopmargin=6pt, innerbottommargin=6pt
]{adaptbox}

\newmdenv[
  backgroundcolor=retryred,
  linecolor=borderred,
  linewidth=1.2pt,
  innerleftmargin=10pt, innerrightmargin=10pt,
  innertopmargin=6pt, innerbottommargin=6pt
]{retrybox}

\newcommand{\field}[1]{\texttt{\small #1}}
\newcommand{\placeholder}[1]{\textcolor{gray}{\textit{$\langle$#1$\rangle$}}}

\usepackage{lineno}

\definecolor{darkblue}{rgb}{0, 0, 0.5}
\hypersetup{colorlinks=true, citecolor=darkblue, linkcolor=darkblue, urlcolor=darkblue}

\title{\ourmodel{}: Optimizing Multimodal Reasoning for Multi-Turn Table Question Answering}

\author{\mdseries
Tung Sum Thomas Kwok$^{1,5}$\thanks{Equal contribution.}, 
Xinyu Wang$^{2,5}$, 
Xiaofeng Lin$^{1}$, 
Peng Lu$^{3}$$^{5}$,
Chunhe Wang$^{1}$, \\
Changlun Li$^{4}$,
Hanwei Wu$^{3,5}$,
Nan Tang$^{4}$, Elisa Kreiss$^{1}$\thanks{Corresponding author.}, Guang Cheng$^{1}$\thanks{Corresponding author.}\\
$^{1}$University of California, Los Angeles, 
$^{2}$McGill University,
$^{3}$Université de Montréal\\
$^{4}$The Hong Kong University of Science and Technology (Guangzhou), \\
$^{5}$SimpleWay \\
\texttt{\href{mailto:tk1018@ucla.edu}{tk1018@ucla.edu},
\href{mailto:ekreiss@ucla.edu}{ekreiss@ucla.edu},
\href{mailto:guangcheng@ucla.edu}{guangcheng@ucla.edu}}
}

\newcommand{\ourmodel}{{\textsc{TabQAWorld}{}}}
\newcommand{\tick}{\ding{51}}
\newcommand{\cross}{\ding{55}}
\usepackage{colortbl}
\usepackage[table]{xcolor}

\definecolor{Best}{RGB}{210,240,220}
\definecolor{Second}{RGB}{245,245,200}
\newcommand{\hilite}[1]{\cellcolor{Best}#1}
\newcommand{\hilight}[1]{\cellcolor{Second}#1}
\usepackage{amsmath,amsfonts,bm}

\def\eqref#1{equation~\ref{#1}}
\def\1{\bm{1}}

\DeclareMathAlphabet{\mathsfit}{\encodingdefault}{\sfdefault}{m}{sl}
\SetMathAlphabet{\mathsfit}{bold}{\encodingdefault}{\sfdefault}{bx}{n}

\begin{document}

\ifcolmsubmission
\linenumbers
\fi

\maketitle

\begin{abstract}
 Multimodal reasoning has emerged as a powerful framework for enhancing reasoning capabilities of reasoning models. While multi-turn table reasoning methods have improved reasoning accuracy through tool use and reward modeling, they rely on fixed text serialization for table state readouts. This introduces representation errors in table encoding that significantly accumulate over multiple turns. Such accumulation is alleviated by tabular grounding methods in the expense of inference compute and cost, rendering real world deployment impractical. To address this, we introduce \ourmodel{}, a table reasoning framework that jointly optimizes tabular action through representation and estimation. For representation, \ourmodel{} employs an action-conditioned multimodal selection policy, which dynamically switches between visual and textual representations to maximize table state readout reliability. For estimation, \ourmodel{} optimizes stepwise reasoning trajectory through table metadata including dimension, data types and key values, safely planning trajectory and compressing low-complexity actions to reduce conversation turns and latency. Designed as a training-free framework, %\textcolor{red}{Designed as a unified framework, \ourmodel{} can be deployed as a training-free test-time scaling framework or via reinforcement learning post-training recipe.} 
empirical evaluations show that \ourmodel{} achieves state-of-the-art performance with \textbf{4.87\%} accuracy improvements over baselines, with \textbf{5.42\%} accuracy gain and \textbf{33.35\%} inference latency reduction over static settings, establishing a new standard for reliable and efficient table reasoning. 

\end{abstract}

\section{Introduction}
Tabular reasoning is an emerging focus of language model studies across applications such as numerical analysis~\citep{akhtar-etal-2023-exploring, 10.1145/3616855.3635752,li2025time}, fact checking~\citep{parikh-etal-2020-totto,nan-etal-2022-fetaqa} and question answering~\citep{pasupat-liang-2015-compositional,10.1145/3654979} (TableQA). Unlike free-form text, tables encode information in structured form with rows and columns. While generative AI solutions such as Text2SQL are effective for database management~\citep{li2025alphasql,rewardsql2025}, they are less suitable for semantic interpretation over unstructured or noisy inputs~\citep{liu-etal-2024-rethinking,abhyankar-etal-2025-h}. Prior work has since established that effective reasoning over tables requires both accurate understanding of tabular context and step-by-step logical inference~\citep{wang2024chainoftable,table-reasoning-survey,zou2026tattoo}. To support such multi-step conversation, recent studies incorporate agentic tool use~\citep{wang2024chainoftable,ji2025treeoftable} and test-time inference~\citep{yang2025triples,yang2025causality} to align language models with demands of complex table understanding and reasoning. %Their execution follows a state-transition loop between table and actions. \textcolor{red}{Make this more explicit reason for why do we need multistep reasoning?} %naturally compatible with internal world model reasoning over evolving states~\citep{xing2025critiquesworldmodels,wu2025rlvrworld}:
%\[
%\texttt{Table} \rightarrow \texttt{Action} \rightarrow \texttt{New Table},
%\]

Yet in practice, multi-step table reasoning fails in tracking belief states, i.e., the representation encoding the table and perceived by reasoning model when determining next steps~\citep{yu2026proactagenticlookaheadinteractive}. Recent work attributes this failure to noisy representation encoding of the original table content, which may corrupt intermediate state understanding~\citep{singha2023tabular}. Fixed text serialization such as \texttt{JSON} or \texttt{Markdown} distorts table topology, as cells that are adjacent in a 2D table become distant in a token sequence. This introduces perceptual errors in row and column understanding, hence weakening reliable retrieval within cells~\citep{wang2025needleinatable}. Propagated errors require additional clarification steps, which leads to trajectory drifts and degraded reliability~\citep{laban2026llms}. This further burdens the originally heavy overhead due to multi-turn conversation. Propagated errors require additional clarification steps, which leads to trajectory drifts and degraded reliability~\citep{laban2026llms}. This further burdens the originally heavy overhead due to multi-turn conversation.%Propagation of errors requires additional clarification steps, which further burdens the multi-turn conversation overhead, leading to trajectory drifts hence degraded reliability~\citep{laban2026llms}. 
In order to address these constraints, prior works incorporate stepwise reflection~\citep{ji2025treeoftable} and process reward modeling~\citep{zou2026tattoo, kwok2026enhancingtableqaverifiablereasoning} to monitor reasoning drifts, while others have explored leveraging tabular grounding and state estimation to improve table understanding~\citep{jiang-etal-2023-structgpt,nguyen2025interpretable,zhou2025tablequestionansweringera}. In essence, all these approaches focus on increasing accuracy at the cost of significantly increasing compute and inference time hence dampening real world deployment, urging exploration in integrating efficient inference and stepwise table reasoning.
%Existing works incorporate stepwise reflection~\citep{ji2025treeoftable} and process reward modeling~\citep{zou2026tattoo, kwok2026enhancingtableqaverifiablereasoning} to monitor reasoning drifts, but they lack trajectory-level monitoring and revision, resulting in increased inference latency. 
%actions are conditioned on these intermediate state estimates, whereas early estimation errors propagate through subsequent transitions, causing trajectory drift and compounding mistakes over time. Naively increasing per-step checking reduces some failures~\citep{ji2025treeoftable} but also increases latency and token cost due to repeated verification. PRM-based methods improve step scoring and re-ranking~\citep{kwok2026enhancingtableqaverifiablereasoning,zou2026tattoo}, but they still often lack explicit trajectory-level state modeling and efficient plan revision under partial observations.
\begin{figure}[t]
    \centering
    \includegraphics[width=0.9\linewidth]{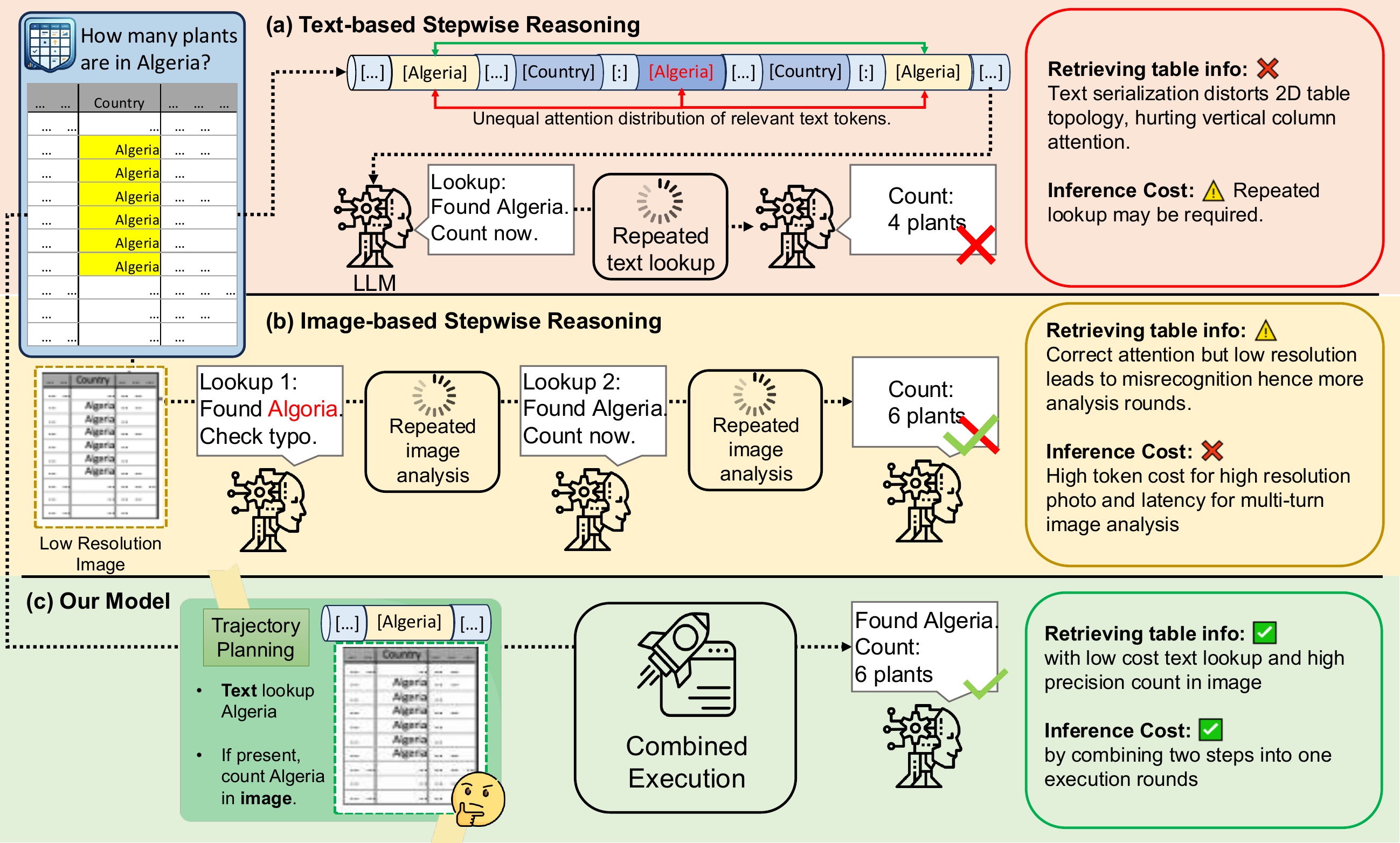}
    \caption{Motivation and overview of \ourmodel{}. Fixed text serialization introduces state tracking noise (representation bottleneck), which propagates across multi-step reasoning and causes trajectory drift (estimation bottleneck). \ourmodel{} addresses such failure process by jointly optimizing \textbf{what to see} and \textbf{what to expect} .}
    \vspace{-\baselineskip}
    \label{fig:motivation}
\end{figure}

To facilitate accurate and efficient table reasoning for practical deployments, %that does not just improve accuracy but also saves compute and inference time for viable application, 
we introduce \ourmodel{}.
Motivated by recent work which argues that the key bottleneck of interactive environment for language-based and long-horizon planning lies on the modality gap~\citep{xu2026visual} and compounding errors when simulating future states~\citep{yu2026proactagenticlookaheadinteractive}, \ourmodel{} jointly optimizes table reasoning actions through \emph{representation}: What to see, and \emph{estimation}: {What to expect}. \ourmodel{} has two components to achieve accurate and efficient table reasoning. (1) An \textbf{action-conditioned multimodal selection policy} optimizes state understanding for next step action: At each operation, the agent adaptively selects the most suitable modality for the table state readout, e.g., through text, vision or multimodal input. (2) \textbf{Metadata-guided trajectory optimization} safely merges low-risk actions for efficient inference: The agent initiates trajectory planning through future state estimation via low-dimensional table metadata. Under this internal state-focused reasoning paradigm, \ourmodel{} is designed for training-free deployment that does not require prior fine-tuning for general table reasoning tasks. %\textcolor{red}{Under this internal state-focused reasoning paradigm, \ourmodel{} is designed as a unified framework capable of training-free test-time scaling and reinforcement learning (RL) post-training deployment.} 

Framed as a training-free framework for multimodal reasoning and trajectory optimizing, \ourmodel{} achieves state-of-the-art performance across existing benchmarks, outperforming previous best methods by 6.32\%, 10.81\% and 4.87\% on the three domains respectively. 
Ablation studies show that both framework components positively contribute to the overall performance gains: action-conditioned adaptive modality achieves 5.42\% accuracy improvement over top-performing fixed modality benchmarks, while metadata-guided optimization contributes to 33.35\% reduction in end-to-end inference latency with comparable compute, indicating more efficient inference. 
%\textcolor{red}{In scenario where \ourmodel{} is used as RL post training, we report additional XX\% accuracy improvement against state-of-the-art baselines.} 
To conclude, we introduce and establish \ourmodel{} in the following three steps:% our contributions are threefold:
%pilot study shows consistent gains from improved perception, with an 8.9\% accuracy improvement on image perception over fixed text serialization, and our proposed adaptive policy adding a further 8.6\%. Our main experiments indicate consistent improvements across four Multimodal Large Language Models (MLLMs) and three TableQA datasets with a 33.18\% reduction in end-to-end inference latency (Section~\ref{sub: ablation}). Additional case study on \ourmodel{} further analyzes how different modalities and question types affect MLLMs' attention on the table (Section~\ref{sub: question-type-vs-modality}). \textcolor{red}{Lastly, \ourmodel{} can be used to post-train an agent via RL, which reports additional XX\% accuracy improvement against state-of-the-art baselines, illustrating the generalizability of \ourmodel{} as both training-free framework and an RL recipe.} To conclude, our contributions are threefold:
\begin{enumerate}
    \item \textbf{Multimodal efficient reasoning for TableQA.} We propose \ourmodel{}, which improves table reasoning through adaptive multimodal representation and trajectory optimization (Section~\ref{sec: methodology}),
    \item \textbf{A plug-and-play, training-free framework for TableQA.} \ourmodel{} augments existing TableQA paradigms with adaptive multimodal and trajectory optimization, improving performance while reducing latency (Section~\ref{sec: methodology}), and
    %\item \textcolor{red}{\textbf{Unified inference and training recipe.} \ourmodel{} supports both plug-and-play test-time scaling and RL-based post-training (Section~\ref{sec: methodology}), and}
    \item \textbf{State-of-the-art TableQA performance .} \ourmodel{} consistently outperforms existing baselines over multiple TableQA datasets and base models (Section~\ref{sec: experiments}).
\end{enumerate}

\section{Related Work}
\subsection{TableQA and Reasoning}
TableQA requires models to jointly understand natural language questions and structured table content. Early approaches focused on semantic parsing, table pre-training, and schema-aware encoders~\citep{herzig-etal-2020-tapas,yin-etal-2020-tabert,jiang2022omnitab,lee-etal-2025-dcg}. Recent LLM-based methods expand this line with prompting, in-context learning, retrieval, and tool use~\citep{cheng2023binding,chen-2023-large,sui-etal-2024-tap4llm,wang2024chainoftable,yang2025triples,yang2025causality,lin-etal-2023-inner,wu-etal-2023-tacr}. Post-training methods, including supervised fine-tuning and reinforcement learning~\citep{zha2023tablegptunifyingtablesnature,zhang2024tablellamaopenlargegeneralist,zou2026tattoo,wu2025tabler1regionbasedreinforcementlearning,yang2025tablegptr1advancingtabularreasoning,guo2026rethinkingtablepruningtableqa}, further improve reasoning quality. In parallel, Multimodal Large Language Models (MLLMs) %\peng{explain MLLM in full before use it}
reasoning studies show that representation format strongly affects downstream reasoning quality~\citep{deng-etal-2024-tables,zhang2024multimodal,lei-etal-2025-scaffolding,hu2024visual,zhou2024imageofthoughtpromptingvisualreasoning,openai2025o3o4mini}, specifically using text-based reasoning in `vision-first' reasoning~\citep{xu2026visual}. This motivates using adaptive modality for TableQA~\citep{xing2026tabledart}. Our work builds on this observation by extending modality choice conditioning on first-class action during reasoning, rather than fixing one static serialization for the full trajectory. 

\subsection{Multi-step Planning for Table Reasoning}
Multi-step reasoning emphasizes explicit state estimation and transition modeling across multi-step interaction~\citep{xing2025critiquesworldmodels,wu2025rlvrworld}. In TableQA, related ideas appear in tabular grounding and self-verification~\citep{wang2024chainoftable,qu2025tabicl,jiang-etal-2023-structgpt,nguyen2025interpretable,ji2025treeoftable}. Step-level reward signals also improve intermediate-step selection~\citep{liu-etal-2024-rethinking,kwok2026enhancingtableqaverifiablereasoning,zou2026tattoo}, but they lack long-horizon trajectory tracking. Action-conditional generative models learn dynamics by predicting future observations conditioned on actions~\citep{world_model,deepmind_genie,hafner2019planet} and optimizing trajectory through simulation~\citep{yu2026proactagenticlookaheadinteractive}, but such adaptation into table reasoning remains underexplored. This motivates exploration of trajectory optimization and efficient LLM inference in complex table reasoning scenarios. 

\section{Preliminary} \label{sec:problem-formulation}
\paragraph{Table Reasoning with MLLMs} \label{sub:state-transition-formulation}
We denote a table as $T=(H,R)$, where $H$ is the set of column headers and $R$ is the set of rows aligned with $H$. Given an initial table $T_0=T$, a natural-language question $q$, and ground-truth answer $\alpha$, the goal is to predict $\hat{\alpha}$ that maximizes a task metric $G(\hat{\alpha},\alpha)$. In multi-turn TableQA, an agent typically cannot process the entire table at every step due to context and efficiency constraints~\citep{wang2025needleinatable}. We therefore model reasoning as a trajectory policy over partial observations, i.e. a partially observable Markov Decision Proces (POMDP). At step $t$, we denote $T_t$ the current table state (an instantiated table after $t{-}1$ edits), $o_t$ the partial observation exposed to the agent under the chosen modality $m$, and $z_t$ the agent's internal belief state carried across steps. Given the initial state $T_0=T$, let $\tau=\{a_1,\ldots,a_S\}$ be a reasoning trajectory, where $a_1,\ldots,a_{S-1}$ are intermediate reasoning steps and $a_S=\hat{\alpha}$ is the final-answer action. The model induces a conditional policy $\pi(\tau\mid T_0,q)$ that interleaves textual reasoning with executable table operations. This formulation highlights two core challenges: (i) selecting actions that remain consistent with that belief~\citep{kim2024openvlaopensourcevisionlanguageactionmodel}, and (ii) maintaining a faithful internal belief over evolving table states~\citep{wang2025vagen,xing2025critiquesworldmodels}.

\section{Table Reasoning via Internal State Belief Models} \label{sec: methodology}
The two core challenges of table reasoning formulation suggests the need of internal state belief modeling, where the MLLM must understand the current table state to act properly, predict how state changes after each action to revise future steps accordingly. To motivate this design, we conduct a pilot study on one key question:

\begin{tcolorbox}[width=0.98\textwidth, colback=blue!5!white, colframe=blue!75!black, boxsep=1pt, top=3pt, bottom=3pt, left=4pt, right=4pt]
    RQ: To what extent does table representation affect information accessibility?
\end{tcolorbox}
Prior work shows that different table representations yield different extraction quality~\citep{xu2026efficienttableretrievalunderstanding,wang2025needleinatable}. We extend this to stepwise reasoning quality by comparing text-based serializations, i.e. \texttt{json}, \texttt{latex}, and \texttt{markdown}, with vision-included parsing, i.e. an image view (table rendered as \texttt{jpg}), and a multimodal view containing image and text headers. We analyze using Qwen-3-8B-VL~\citep{bai2025qwen3vltechnicalreport} on three complex reasoning datasets, namely WikiTableQuestions (WTQ)~\citep{pasupat-liang-2015-compositional}, MMQA~\citep{wu2025mmqa}, and MMTU~\citep{xing2025mmtu}, with the latter two datasets released in 2025 to simulate zero-shot environment without pre-training bias. Table~\ref{tab:multimodal_perception} shows a consistent accuracy improvement and compute reduction with vision-included representations. %Although visual tokens can be more costly per token in proprietary pricing schemes~\citep{openai_api_pricing,google_gemini_api_pricing}, improved state readability shortens trajectories, hence lowering overall inference overhead in both token costs and conversation turns. 
\begin{tcolorbox}[width=0.98\textwidth, colback=yellow!5!white, colframe=yellow!75!black, boxsep=1pt, top=3pt, bottom=3pt, left=4pt, right=4pt]
    \textbf{Observation (Importance of Internal State for Table Reasoning)}: Agents perceiving vision-based tables outperforms with both higher accuracy and lower inference cost. 
\end{tcolorbox}
\begin{wraptable}{r}{0.5\textwidth}
\vspace{-\baselineskip}
\centering
\vspace{-\baselineskip}
\caption{Reasoning performance and costs.}
\label{tab:multimodal_perception}
\setlength{\tabcolsep}{2pt} % Tighten column spacing
\resizebox{0.95\linewidth}{!}{%
\begin{tabular}{lccc}
\toprule
 & WTQ & MMQA & MMTU \\
\midrule
\multicolumn{4}{l}{\textit{Accuracy $\pm$ SD under different perceived modality}} \\
\texttt{json}     & 0.722 {\small{(0.032)}} & 0.594 {\small{(0.035)}} & 0.303 {\small{(0.033)}}\\
\texttt{latex}    & 0.707 {\small{(0.032)}} & 0.574 {\small{(0.035)}} & 0.333 {\small{(0.034)}}\\
\texttt{Markdown} & 0.712 {\small{(0.032)}} & 0.584 {\small{(0.035)}} & 0.364 {\small{(0.034)}}\\
Image             & \textbf{0.818} {\small{\textbf{(0.027)}}} & 0.614 {\small{(0.035)}} & \textbf{0.374} {\small{(0.034)}}\\
Multimodal        & 0.788 {\small{(0.029)}} & \textbf{0.635} {\small{\textbf{(0.034)}}} & 0.364 {\small{(0.034)}}\\
\midrule 
\multicolumn{4}{l}{\textit{Accuracy $\pm$ SD under different estimated state}} \\
No estimate & \textbf{0.818} {\small{\textbf{(0.027)}}} & 0.614 {\small{(0.035)}} & \textbf{0.374} {\small{(0.034)}}\\
Table estimate & {0.753} {\small{{(0.031)}}} & 0.543 {\small{(0.036)}} & {0.303} {\small{(0.033)}}\\
\midrule
\multicolumn{4}{l}{\textit{Token costs (in $10^{3}$ tokens) / Conversation Turns}} \\
\texttt{json} & 14.6 / {\small{8.83}} & 7.4 / {\small 5.13} & 23.5 / {\small 3.55} \\
\texttt{latex}    & 12.6 / {\small 9.53} & 6.1 / {\small 5.20} & 27.5 / {\small 3.41} \\
\texttt{Markdown} & 11.8 / {\small 9.01} & 6.5 / {\small 4.54} & 17.5 / {\small 4.16} \\
Image             & 8.4 / {\small 5.58}  & \textbf{5.3} / {\small 2.86} & \textbf{16.8} / {\small 2.16} \\
Multimodal        & \textbf{6.0} / {\small 4.66} & 5.8 / {\small 2.41} & 17.0 / {\small 2.51} \\
\bottomrule
\end{tabular}
}
\vspace{-\baselineskip}
\end{wraptable}
\subsection{Performance Analysis.} 
\begin{figure}[H]
    
    \centering
    %\vspace{-\baselineskip}
    \includegraphics[width=0.95\linewidth]{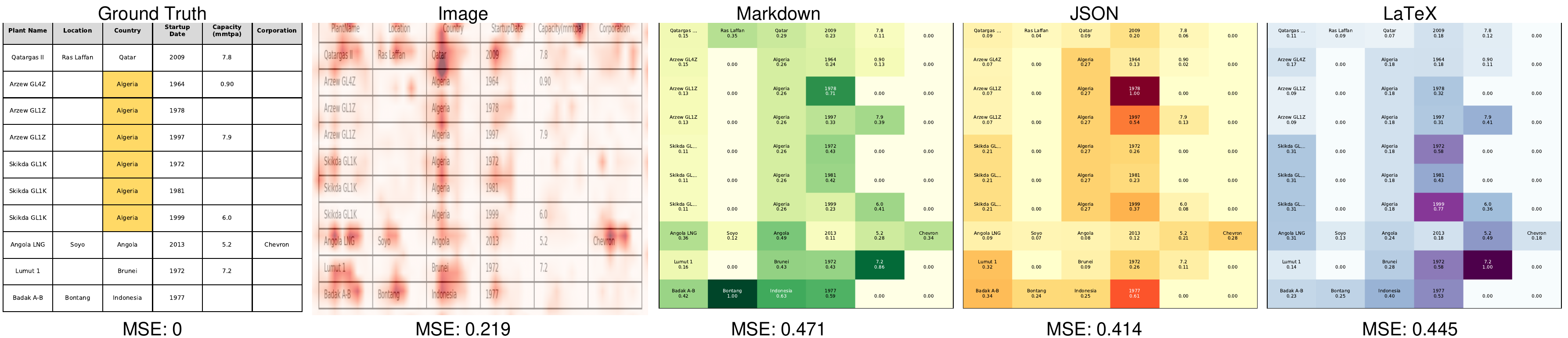}
    \caption{An illustrative example of how image-based parsing facilitates more human-preferred column attention than text-serialized tables. The value below each table indicates the mean-squared error (MSE) against human-preferred binary attention.}
\label{fig:atten-illus}
\vspace{-\baselineskip}
\end{figure}

This motivates error localization on incorrect trajectories, allowing us to identify two existing table reasoning bottlenecks:
\paragraph{(1) Serializing tables weakens 2D structure awareness for row computation.} The WTQ question ``How many plants are in Algeria?'' requires counting ``Algeria'' occurrence within ``Country'' column. Figure~\ref{fig:atten-illus} shows sharper attention from image-based parsing on relevant entries compared to diffused attention from text serializations. Lower total token usage and fewer turns with visual representations in Table~\ref{tab:multimodal_perception} suggest that state representation impacts information accessibility and prevents drifts in long interactions~\citep{laban2026llms}. This is consistent with prior evidence on preserved information under visual compression~\citep{shi2026codeocreffectivenessvisionlanguage}. We include detailed case studies on how agent attends to different table representations in  Appendix~\ref{sub: table-representation-attention} and ~\ref{sec:text-serialization-case-study}. 
\paragraph{(2) Full table state estimation is challenging for frontier MLLMs.}
While modeling belief state is considered a better objective for planning under POMDP assumption~\citep{10.5555/1643275.1643301}, Table \ref{tab:multimodal_perception} shows degraded accuracy when MLLM incorporates table estimates to assist each reasoning step. This suggests that predicting full table is challenging even to frontier LLMs due to the complexity and fragility of structured data. In Figure \ref{fig:gpt-full-table-hallucination}, we ask GPT-5.4~\citep{openai2026gpt54} to estimate table state after sorting the table. While the order is sorted correctly, we observe LLM hallucinates and generates `1' for all numerical values.

\begin{figure}[t]
\centering

\begin{minipage}[t]{0.44\textwidth}
\vspace{0pt}
\begin{tcolorbox}[
    colback=yellow!5!white,
    colframe=yellow!75!black,
    boxsep=1pt,
    top=3pt,
    bottom=3pt,
    left=4pt,
    right=4pt
]
\textbf{Takeaways (Key Bottlenecks of Multi-turn TableQA).}

\begin{itemize}[noitemsep,topsep=2pt,leftmargin=1.2em]
\item MLLMs are sensitive to representation, where visual grounding improves information accessibility and results in higher answer accuracy and fewer interaction rounds.
\item Belief state estimation for structured table is fragile to LLM hallucinations.
\end{itemize}
\end{tcolorbox}
\end{minipage}
\hfill
\begin{minipage}[t]{0.52\textwidth}
\vspace{0pt}
\centering
\setlength{\tabcolsep}{3pt}
\renewcommand{\arraystretch}{0.95}
\small

\begin{subfigure}[t]{0.32\linewidth}
\centering
\begin{tabular}{l}
\toprule
(a) Original \\
\midrule
E. Keene: 2 \\
P. Norris: 3 \\
K. Biller: 2 \\
M. Fresco: 2 \\
others (11): 1 \\
\bottomrule
\end{tabular}
\end{subfigure}
\hfill
\begin{subfigure}[t]{0.32\linewidth}
\centering
\begin{tabular}{l}
\toprule
(b) Truth \\
\midrule 
P. Norris: 3 \\
E. Keene: 2 \\
K. Biller: 2 \\
M. Fresco: 2 \\
others (11): 1 \\
\bottomrule
\end{tabular}
\end{subfigure}
\hfill
\begin{subfigure}[t]{0.32\linewidth}
\centering
\begin{tabular}{l}
\toprule
(c) GPT-5.4 \\
\midrule 
P. Norris: \textcolor{red}{1} \\
E. Keene: \textcolor{red}{1} \\
K. Biller: \textcolor{red}{1} \\
M. Fresco: \textcolor{red}{1} \\
others (11): 1 \\
\bottomrule
\end{tabular}
\end{subfigure}

\caption{Hallucinations in full table estimation from frontier GPT-5.4 motivate lower-dimensional state estimation.}
\label{fig:gpt-full-table-hallucination}
\end{minipage}

\end{figure}

These findings motivate \ourmodel{} with (i) action-conditioned multimodal selection and (ii) metadata-guided trajectory optimization to improve table representation and estimation.

\begin{figure}[H]
    \centering
    \includegraphics[width=0.95\linewidth]{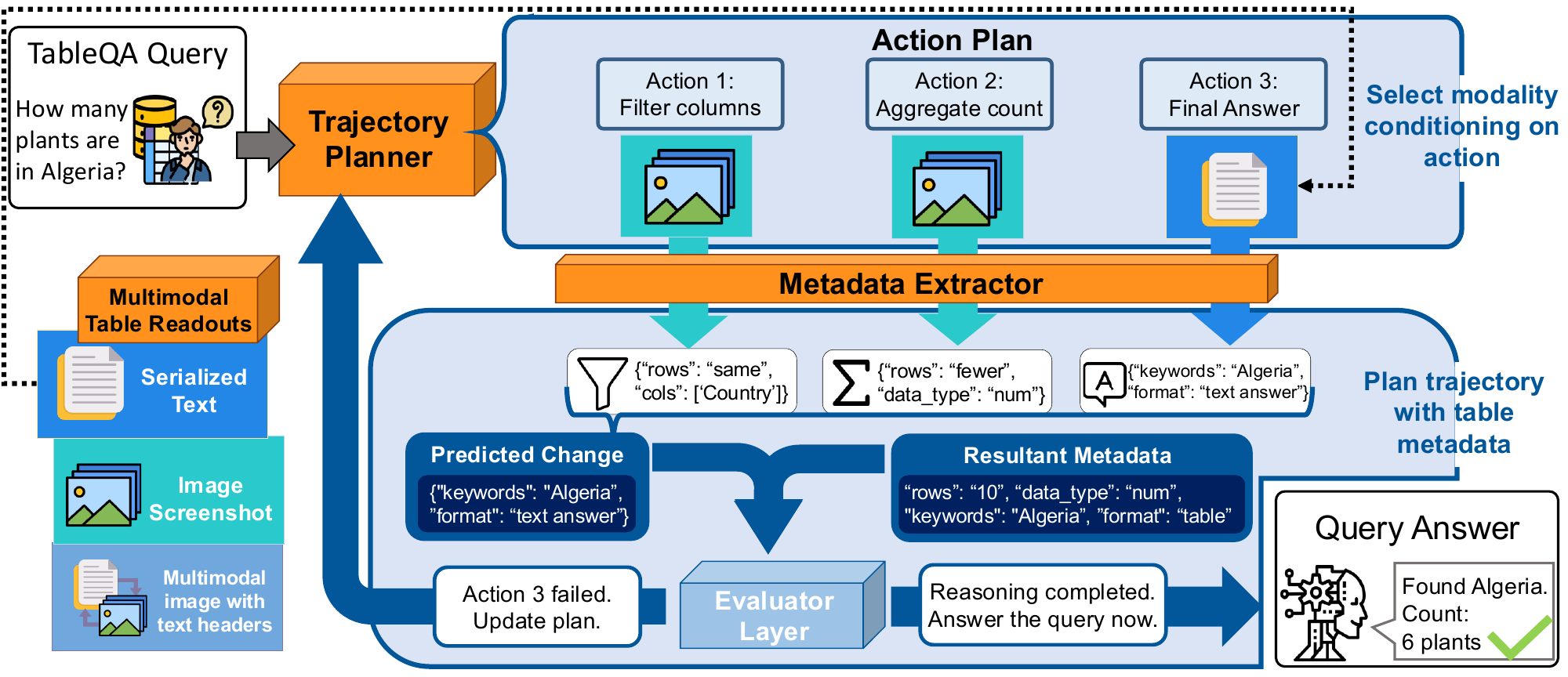}
    \caption{\ourmodel{} dynamically selects the optimal data modality based on task purposes, and optimizes reasoning trajectory based on low dimensional metadata to minimize token usage and latency while maintaining a rigorous feedback loop to ensure convergence on an accurate final answer.}
    %\vspace{-\baselineskip}
    \label{fig:workflow}
\end{figure}

\subsection{Action-conditioned Multimodal Selection Policy}
A key failure mode in multi-turn TableQA is observation mismatch, where the action may be correct in principle, but the chosen representation obscures the evidence needed for reliable execution. \ourmodel{} addresses this by selecting \emph{what to observe next}. Let $g(\cdot;m)$ be a modality-conditioned encoder with $m\in\mathcal{M}$. At step $t$, the agent predicts a triplet $a_t=\{\alpha_t,r_t,m_t\}$, where $\alpha_t$ is the executable operation, $r_t$ is step-level feedback, and $m_t$ is the modality used to encode the next observation.
\begin{align*}
x_t \mapsto a_t=\{\alpha_t,r_t,m_t\}, \text{where } 
T_{t+1} = a_t(T_t),\ o_{t+1}=\Omega(T_{t+1}),\
x_{t+1} = g(o_{t+1};m_t).
\end{align*}
This formulation turns modality into a decision policy rather than a fixed input format. In practice, the policy offers MLLM the flexibility to dynamically select the preferred representation, in order to improve information representativeness under a fixed context budget. As a result, later decisions are conditioned on more faithful state encoding, which reduces downstream error propagation across long trajectories.

\subsection{Metadata-guided Trajectory Optimization}
% Start the wrap: {r} for right side, 0.5\textwidth for width
%\begin{wrapfigure}{r}{0.99\textwidth}
\begin{figure}[t]
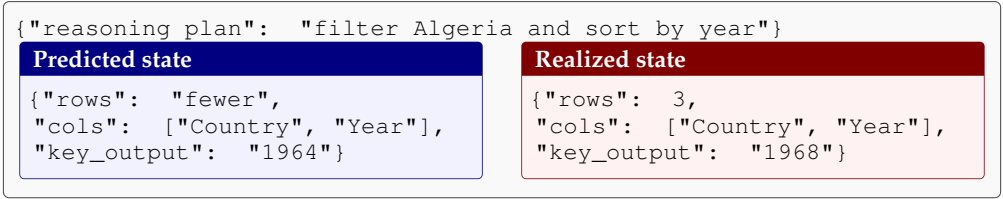

\centering
\small
\begin{minipage}{0.95\textwidth}
\begin{tcolorbox}[
    colback=gray!5,
    colframe=black!70,
    boxrule=0.4pt,
    arc=2pt,
    left=3pt,right=3pt,top=4pt,bottom=4pt
]
\texttt{\{"reasoning plan": "filter Algeria and sort by year"\}}

\begin{minipage}[t]{0.48\textwidth}
\begin{tcolorbox}[
    colback=blue!5,
    colframe=blue!50!black,
    boxrule=0.3pt,
    arc=1.5pt,
    left=2pt,right=2pt,top=3pt,bottom=3pt,
    title=\textbf{Predicted state}
]
\texttt{\{"rows": "fewer",}\\
\texttt{"cols": ["Country", "Year"],}\\
\texttt{"key\_output": "1964"\}}
\end{tcolorbox}
\end{minipage}
\hfill
\begin{minipage}[t]{0.48\textwidth}
\begin{tcolorbox}[
    colback=red!5,
    colframe=red!50!black,
    boxrule=0.3pt,
    arc=1.5pt,
    left=2pt,right=2pt,top=3pt,bottom=3pt,
    title=\textbf{Realized state}
]
\texttt{\{"rows": 3,}\\
\texttt{"cols": ["Country", "Year"],}\\
\texttt{"key\_output": "1968"\}}
\end{tcolorbox}
\end{minipage}
\end{tcolorbox}
\end{minipage}
\caption{Illustration of metadata-guided execution. A mismatch in \texttt{key\_output} (1964 vs.\ 1968) triggers replanning.}
\label{fig:predicted-metadata-json}
\end{figure}

As table state estimation is infeasible (Figure~\ref{fig:gpt-full-table-hallucination}), we propose using a lightweight metadata summary of the table state as a low-dimension tabular projection (Figure~\ref{fig:predicted-metadata-json}). This serves as a state estimation shortcut to facilitate trajectory planning and optimization. Nevertheless, even with simplified state estimation, strict step-by-step execution still incurs large interaction overhead and can amplify drift when plan becomes long. To address this, we optimize the execution trajectory by compressing low-risk actions with online verification. At each execution, the agent maintains a current plan and expected metadata states, and revises the plan if expected metadata misaligns with realized state. Formally, we denote $z_t=\phi(T_t)$ as a low-dimensional metadata of table state $T_t$. At each step $t$, the agent predicts $\hat{z}_{t+1}$ and compares against the realized state $z_{t+1}$. Mismatch between predicted and realized states reflects execution uncertainty, which then guides reasoning plan updates and compression of low uncertainty actions only. 
\begin{align*}
(z_t,\hat{z}_{t+1},z_{t+1})&=\bigl(\phi(T_t),\phi(\hat{T}_{t+1}),\phi(T_{t+1})\bigr).,\quad
c(a_t)=\nabla(z_t,\hat{z}_{t+1}),\quad \nu(a_t)=\nabla(z_{t+1},\hat{z}_{t+1}).
%\text{where }\tilde{a}_j &= a_{v_j}\circ \cdots \circ a_{u_j}, \text{ such that }\sum_{k=u_j}^{v_j} c(a_k)\le \gamma \;\text{and}\; \max_{k\in[u_j,v_j]} \nu(a_k)\le \delta.
\end{align*}
%where $c(a_t)$ is estimated transition complexity, $\nu(a_t)$ is transition uncertainty measured by predicted--observed state mismatch, $\gamma$ is the compression budget, and $\delta$ is a safety threshold. Intuitively, only low-complexity and low-uncertainty segments are merged; high-risk transitions remain fine-grained for controllability and recovery. This yields a plan--verify--revise loop that improves efficiency without sacrificing robustness.
\section{Experimental Setup} \label{sec: experiments}
%We evaluate \ourmodel{} in two regimes: (i) training-free plug-and-play inference and \textcolor{red}{(ii) RL-based post-training}. Experiments are designed to answer two questions: whether the action-conditioned multimodal selection policy and metadata-guided trajectory optimization improve accuracy and latency respectively, and whether these gains transfer across model families.
%\subsection{Experimental Setup} \label{sub: experiment-setup}
\paragraph{Evaluation Datasets and Metrics.} We evaluate \ourmodel{} by following established table reasoning evaluation protocol from \cite{zheng-etal-2024-multimodal,wang2026hippoenhancingtableunderstanding} on seven diverse benchmarks across TableQA and Table Fact Verification (TFV) tasks. For TableQA which requires interpreting complex queries~\citep{10.1145/3394592,10.1145/3404835.3462839}, we evaluate on five benchmarks, namely WTQ~\citep{pasupat-liang-2015-compositional}, TABMWP~\citep{lu2023dynamic}, TAT-QA~\citep{zhu-etal-2021-tat}, HiTab~\citep{cheng-etal-2022-hitab}, and FeTaQA~\citep{nan-etal-2022-fetaqa}. For TFV tasks for grounded reasoning, we use TabFact~\citep{Chen2020TabFact:}, InfoTabs~\citep{gupta-etal-2020-infotabs}. Regarding evaluation metrics, standard protocol uses BLEU score~\citep{10.3115/1073083.1073135} on FeTaQA for its free-form response task and binary accuracy for the remaining six datasets. 

\paragraph{Baselines.} We compare \ourmodel{} against a comprehensive set of baselines, including (1) Table-as-Text models containing Llama-2-7B~\citep{touvron2023llama2openfoundation}, Llama3-Instruct-8B~\citep{grattafiori2024llama3herdmodels}, TableLlama-7B~\citep{zhang-etal-2024-tablellama}, (2) Table-as-Image models with Table-LLaVA-7B~\citep{zheng-etal-2024-multimodal}, SynTab-LLaVA-7B~\citep{11093154}, MiniCPM-V-2.6-8B~\citep{minicpm}, Qwen2.5-VL-7B~\citep{bai2025qwen25vltechnicalreport} and Qwen3-VL-8B~\citep{bai2025qwen3vltechnicalreport}, (3) Table-as-Multimodality Baselines with HIPPO-8B~\citep{wang2026hippoenhancingtableunderstanding} and Google Gemini 2.0 Flash~\citep{comanici2025gemini25pushingfrontier}, (4) Training-free TableQA agents using proprietary GPT3.5, including DATER~\citep{10.1145/3539618.3591708}, ReAcTable~\citep{10.14778/3659437.3659452}, Mix-SC~\citep{liu-etal-2024-rethinking}, TIDE~\citep{yang2025triples} and CIT-DP~\citep{yang2025causality}, (5) Adaptive modality TableDART~\citep{xing2026tabledart} integrating TableGPT2-7B~\citep{su2024tablegpt2largemultimodalmodel} with adaptive router fine-tuned from Qwen2.5-VL-7B and Ovis2-8B~\citep{lu2024ovisstructuralembeddingalignment}, (6) Trained agents on trajectory optimization including PRM models~\citep{zou2026tattoo} incorporated on DeepSeek-R1-Distilled-Qwen-14B~\citep{Guo_2025}, and table pruning TabTrim~\citep{guo2026rethinkingtablepruningtableqa} trained on Qwen3 family~\citep{yang2025qwen3technicalreport}. To compare fairly with each baseline subset, we employ Qwen2.5-VL-7B and Qwen3-VL-8B to match model usage in TableDART and TabTrim respectively. For subgroups (4) and (6), state-of-the-art representations CIT-DP, TATTOO and TabTrim are unavailable for reproduction, so that we report as-is accuracy from the published works. Appendix~\ref{app:comparative-models} provides a detailed motivation for benchmark selection and clarifies sourcing of baseline results in previous studies. 

\paragraph{Implementation Details.} To test for generalisability, we incorporate a standardized prompt for all training datasets to avoid inductive bias on datasets. We conduct local inference with Ollama~\citep{ollama2025} for the two open-source model, with hyperparameters following protocol of prior works~\citep{zou2026tattoo,xing2026tabledart}. The complete details regarding prompts, data construction, hyperparameter settings, and computational environment are provided in Appendix~\ref{sec: implementation-details}. 

\section{Main Results}
\begin{table}[htbp]
\centering
\caption{Benchmark results for \ourmodel{} and baseline models.}
\label{tab:main-results}
\resizebox{\textwidth}{!}{
\begin{tabular}{lccccccccc}
\toprule
 & \multicolumn{5}{c}{\textbf{TQA}} & \multicolumn{2}{c}{\textbf{TFV}} & \multicolumn{2}{c}{\textbf{Summary}} \\
\cmidrule(lr){2-6} \cmidrule(lr){7-8} \cmidrule(lr){9-10}
\textbf{Method} & WTQ & TABMWP & TAT-QA & HiTab & FeTaQA & TabFact & InfoTabs & WTQ + TabFact & Average \\
 & (Acc.) & (Acc.) & (Acc.) & (Acc.) & (BLEU) & (Acc.) & (Acc.) & (Acc.) & (Acc.) \\
\midrule
\rowcolor{gray!10} \multicolumn{10}{l}{\textit{Table-as-Text Baselines}} \\
Llama-2-7B & 16.39 & 22.82 & 13.73 & 10.72 & 10.93 & 9.20 & 38.92 & 12.78 & 18.63 \\
Llama3-Instruct-8B & 21.24 & 42.01 & 13.08 & 6.97 & 12.66 & 73.89 & 54.00 & 47.57 & 35.20 \\
TableLlama-7B & 24.97 & 10.10 & 19.04 & 46.57 & {38.38} & 79.37 & 46.57 & 52.17 & 37.77 \\
\midrule
\rowcolor{gray!10} \multicolumn{10}{l}{\textit{Table-as-Image Baselines}} \\
%MiniGPT-4-7B & 0.90 & 0.22 & 0.13 & 0.20 & 0.39 & 0.00 & 0.10 & 0.26 \\
%mPLUG-Owl-7B & 0.62 & 1.76 & 0.13 & 0.25 & 7.42 & 7.46 & 5.53 & 2.63 \\
%mPLUG-Owl2-7B & 0.67 & 6.83 & 0.39 & 0.13 & 11.91 & 8.21 & 26.19 & 7.07 \\
%LLaVA v1.5 & 1.24 & 6.05 & 2.97 & 2.03 & 8.24 & 18.90 & 28.31 & 9.92 \\
Table-LLaVA-7B & 18.43 & 57.78 & 12.82 & 10.09 & 25.60 & 59.85 & 65.26 & 41.85 & 37.37 \\
%Qwen-VL-7B & 0.09 & 3.30 & 0.13 & 0.06 & 0.45 & 1.12 & 0.65 & 0.89 \\
%InternLM-XComposer2-7B & 0.05 & 0.06 & 0.26 & 0.12 & 2.62 & 1.19 & 1.11 & 0.46 \\
%Monkey-7B & 19.07 & 13.26 & 12.31 & 6.41 & 3.41 & 22.56 & 22.11 & 15.95 \\
%TabPedia-7B & 23.53 & 10.66 & 13.08 & 6.54 & 14.31 & 35.49 & 2.43 & 15.29 \\
SynTab-LLaVA-7B & 39.59 & \underline{88.30} & 51.94 & 35.66 & 35.45 & 70.78 & 69.42 & 54.51 & 59.28 \\
MiniCPM-V-2.6-8B & 47.97 & 83.68 & 51.55 & 56.53 & 32.68 & 78.48 & 73.03 & 63.23 & 65.21 \\
Qwen2.5-VL-7B & 54.37 & 63.69 & 51.94 & 62.69 & 10.99 & 75.81 & 70.13 & 65.09 & 63.11 \\
Qwen3-VL-8B & \underline{81.82} & 79.90 & 59.80 & 65.83 & 13.89 & 82.41 & 75.88 & 82.12 & 73.77 \\
\midrule
\rowcolor{gray!10} \multicolumn{10}{l}{\textit{Table-as-Multimodality Baselines}} \\
HIPPO-8B & 55.77 & {87.50} & {60.75} & 63.00 & 33.18 & {82.27} & {75.74} & 69.02 & {70.84} \\
Gemini 2.0 Flash & 63.56 & 46.29 & 35.62 & 60.41 & 10.57 & 81.33 & 54.31 & 72.45 & 56.92 \\
\midrule
\rowcolor{gray!10} \multicolumn{10}{l}{\textit{Training-free TableQA Agent Baselines (Proprietary GPT3.5)}} \\
DATER & 65.9 & —  & — & — & 30.92 & 85.60& — & 75.75& —\\
ReAcTable & 68.0 & — & — & — & 30.43 & 86.10& — & 77.10& —\\
Mix-SC& 73.7 & — & — & — & — & 88.50& — & 81.10& —\\
TIDE & 75.0 & — & — & — & — & 89.82 & — & 82.41& —\\
{CIT-DP} & {76.4} & — & — & — &\textbf{36.34} & \underline{91.30} & — & 83.85& —\\
\midrule
\rowcolor{gray!10} \multicolumn{10}{l}{\textit{Dynamic Adaptive Routing incorporated on TableGPT2-7B (TG2-7B) Base Model)}} \\
TableGPT2-7B (\textit{Text-only Path}) & 61.42 & 83.87 & 50.39 & 70.27 & 28.97 & 77.80 & 71.07 & 69.61 & 69.14 \\
Ovis2-8B (\textit{Image-only Path}) & 58.76 & 87.00 & 47.67 & 68.59 & 34.70 & 80.80 & 74.11 & 70.28 & 69.49 \\
TableDART (TG2-7B+Qwen2.5-VL-7B) & {69.29} & 72.61 & 59.07 & {71.13} & 29.87 & 77.94 & 71.46 & 73.62 & 70.25 \\
{TableDART (TG2-7B+Ovis2-8B)} & {70.58} & 84.54 & {62.05} & {74.37} & \underline{36.11} & {81.37} & {76.22} & 75.98 & {74.86} \\
\midrule
\rowcolor{gray!10} \multicolumn{10}{l}{ \textit{Trained trajectory optimizing agents with (1) PRM in DeepSeek-R1-Distilled-Qwen-14B (DS) and (2) table pruning models trained from Qwen3 model family}} \\
DS + Qwen2.5-Math-PRM-72B & 69.20 & —& —& —& —& 57.90 & — & 63.55 & —\\
DS-Q14B + TATTOO (Qwen3-8B) & 69.80 & —& —& —& —& 58.79 & — & 64.30 & —\\
TabTrim-4B & 76.80 & —& —& —& —& 89.40 & — & 83.10& —\\
TabTrim-8B & 79.40 & —& —& —& —& 91.20 & — & \underline{85.30} & —\\
\midrule
\rowcolor{gray!10} \multicolumn{10}{l}{\textit{\ourmodel{} (Training-free)}} \\
Qwen2.5-VL-7B + \ourmodel{} & {70.20}  & 86.43 & \underline{71.36} & \underline{76.88} & 20.47 & 84.42 & \underline{76.38} & 77.31 & \underline{77.61}\\
\textbf{Qwen3-VL-8B} + \ourmodel{} & \textbf{88.89} & \textbf{94.97} & \textbf{75.88} & \textbf{81.41} & 31.78 & \textbf{91.45} & \textbf{81.41} & \textbf{90.17} & \textbf{85.67}\\
\bottomrule
\end{tabular}
}
\end{table}

\paragraph{State-of-the-Art Performance} Table~\ref{tab:main-results} shows the performance of \ourmodel{} compared to baselines. We observe that \ourmodel{} achieves consistent outperformance, including state-of-the-art works that focus on adaptive modality and trajectory optimization components of \ourmodel{} under a training-free setting, demonstrating that the framework successfully guides MLLMs to leverage their pre-trained knowledge to outperform under efficient reasoning paradigm alone. This resonates with existing visual reasoning work~\citep{xu2026visual} where modality serves as the key bottleneck of table understanding instead of training techniques. Specifically, in the `Average' column, \ourmodel{} achieves the strongest results among all baselines with both backbone models, surpassing the best dynamic adaptive routing model TableDART with TableGPT2-7B and Ovis2-8B by a decisive 7.36\% under the same Qwen2.5-VL-7B setting. \ourmodel{}, under its training-free setting on existing MLLMs, outperforms its backbone model in BLEU score, with comparable performance as trained TableQA agents. This validates \ourmodel{}'s model-agnostic effectiveness.%, \textcolor{red}{where our additional post-training with RL can address the format-focus issue of BLEU score and immediately results in competitive performance against state-of-the-art baselines}.  

\paragraph{Generalization Performance on Component-specific Baselines} \ourmodel{}'s three key component focus on training-free, adaptable modality and trajectory optimization, which it outperforms existing works under each of the three groups. (1) Given the training-free setup, \ourmodel{} incorporated with Qwen3-VL-8B outperforms all five TableQA agent baselines incorporated on proprietary GPT3.5 model in terms of answer accuracy by at least 6.32\%, with competitive performance on BLEU score for FeTaQA. (2) By comparing with dynamic modality baselines, under the same base model, \ourmodel{} shows higher table understanding capability than TableDART, outperforming it on six out of the seven benchmarks by 10.81\% on average. (3) By comparing with works that optimize trajectories where open-source access is unavailable, \ourmodel{} consistently beats all baselines in reported results of both benchmarks under the same generation Qwen3 family by 4.87\% in average, while achieving comparable performance even when using a downgraded Qwen2.5-VL family, confirming that the performance gains are driven by \ourmodel{} intelligent action optimization mechanisms instead of the capacity of backbone models.

%We conduct initial comparison with existing works by referencing to the results in \citep{yang2025causality,zou2026tattoo}. Specifically, we conduct a Pass@4 inference on the datasets utilized in the two works for fair comparison. Our first observation is that by replacing the text-only Qwen3-8B with its MLLM counterpart Qwen3-VL-8B, the agent already yields improved table reasoning accuracy, even when the table input for both settings is kept in text form. This suggests that the addition of vision encoder with sufficient post-training changes the backbone's learned priors about document structure, hence allowing improved spatial understanding regarding tabular layout. Next, by further leveraging vision capabilities of MLLM and parsing the tables as images, the accuracy further improves to outperform baselines with significantly larger proprietary model. This result motivates the idea where diverse tabular representation modes improve accessibility of tabular information even under a training-free environment. It resonates with recent work on purely image-based reasoning~\citep{xu2026visual}, which outlines an inherent modality gap as language may not be the most effective representation for vision-first problem. These findings hence motivate further exploration of multimodal adaptation for tabular tasks. We then conduct detailed ablation experiments to analyze the effectiveness of the two proposed components respectively. 

\section{Ablation Studies} \label{sub: ablation}
The ablation study follows similar protocol as Section~\ref{sec: experiments} and ~\cite{xing2026tabledart} to evaluate component effectiveness with accuracy, and inference latency (in seconds), adding proprietary GPT-5-nano~\citep{openai_gpt5nano_2025} and GPT-5.4 as additional models. We first evaluate the model-agnostic gains jointly brought by the two components, followed by specific benefits on contribution of each component. Lastly, we evaluate the inference efficiency of \ourmodel{}. Detailed motivation for dataset selection and additional BLUE and ROUGE-L~\citep{rouge} evaluations are provided in Appendix~\ref{app:datasets-and-metrics} and~\ref{sec: main-results-detailed} respectively. 

\begin{figure}[t]
    \centering
    \includegraphics[width=0.95\linewidth]{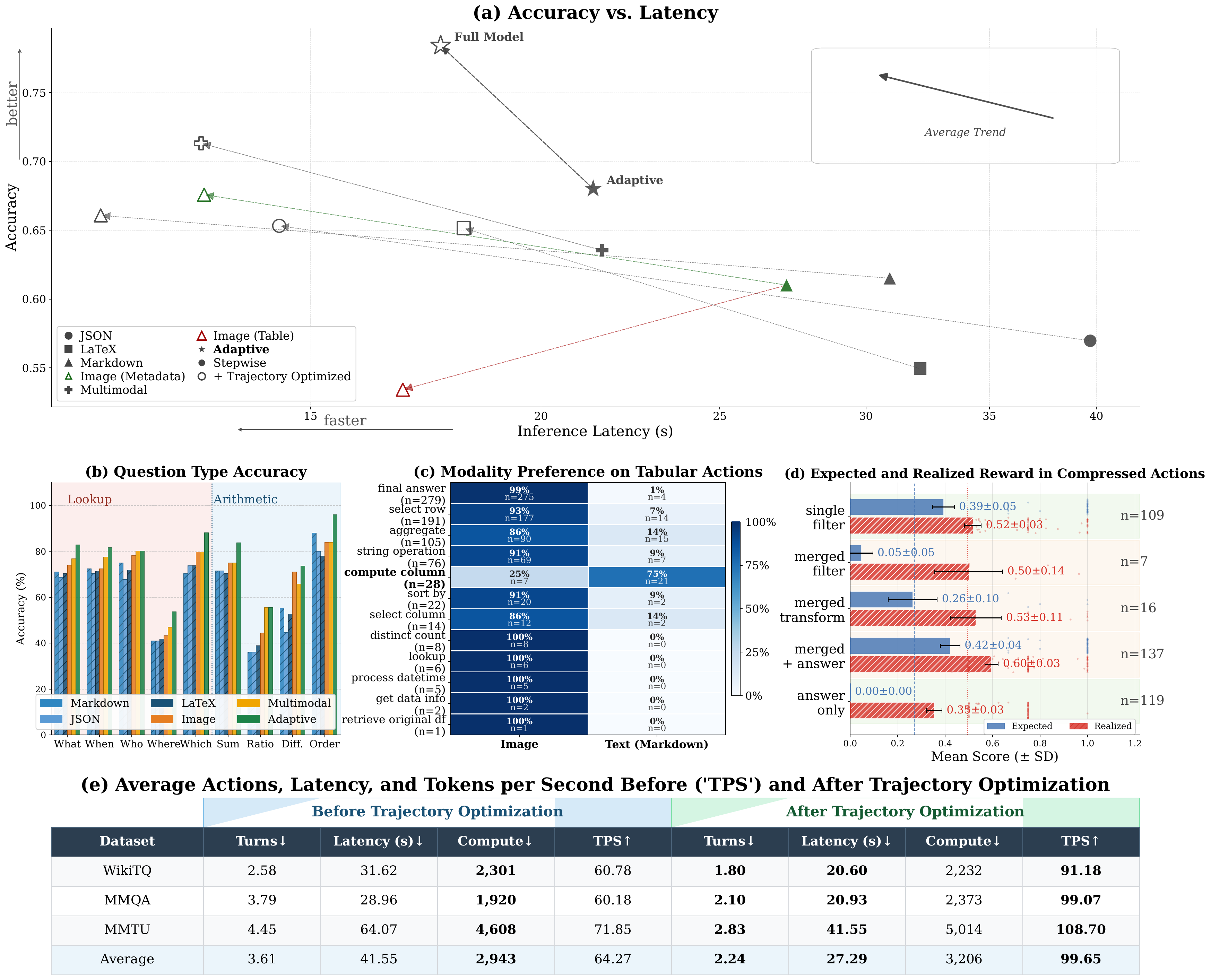}
    \caption{Ablation studies of \ourmodel{}. (a): \ourmodel{} brings model-agnostic joint improvement in accuracy and latency. (b): Visual-based modalities consistently outperform across question types. (c, d): \ourmodel{} agent has internalized respective policy in modality selection and action compression preference. (e): Inference latency after trajectory optimization drops while retaining comparable compute.}
    \label{fig:main-ablation}
    %\vspace{-\baselineskip}
\end{figure}

\subsection{Model-agnostic gains from two components}
Figure~\ref{fig:main-ablation}(a) indicates that across model groups, dynamic modality choice yields an average 5.42\% accuracy improvement over best performing fixed-modality, suggesting how reliable table attention improves tabular information understanding. While visual-included settings report lengthened inference latency, trajectory optimization via metadata effectively reduces average latency time by about 33.35\%, offsetting the incurred latency hence allowing a more efficient tabular agent inference with enhanced reasoning capability. Detailed numerical results are reported in Appendix~\ref{sec: main-results-detailed}. Next, we perform an action-level analysis to understand how \ourmodel{} benefits different tabular actions. 

\subsection{Improved column attention under visual-supported readouts} \label{sub: question-type-vs-modality}
Following~\cite{yang2025causality}, we group questions into nine types spanning lookup and arithmetic behaviors. Figure~\ref{fig:main-ablation}(b) indicates that operations requiring cross-row aggregation improve the most under visual attention. Precisely, vision-supported modalities constantly outperform text-serialized tables in `which' and most arithmetic-based tasks, which require cross-row handling and computation. Nevertheless, all modalities still struggle in arithmetic operations that require particular order, i.e. ratio and difference. This indicates that multi-cell ordered computation still remains to be a table reasoning bottleneck given existing MLLM's reasoning ability.

\subsection{Internalization of modal selection and action compression policy}
Figure~\ref{fig:main-ablation}(c) and (d) show that the agent has internalized policies for both \ourmodel{} components. Firstly, Figure~\ref{fig:main-ablation}(c) shows internalization of a tool-aware action-conditioned multimodal policy instead of relying on fixed representation, even under a training-free setting. While the agent defaults to image modality in most actions, it shows a clear preference for switching to text in \texttt{compute column}, which is a row-computation action with new values derived from computations across columns within each row. Next, Figure~\ref{fig:main-ablation}(d) shows high alignment between expected and realized rewards usually results in high compression preference, i.e. \texttt{single filter} and \texttt{merged + answer} batches, in contrast with \texttt{merged filter} and \texttt{merged transform}. This suggests that not only tabular metadata provides a grounded confidence signal for safe action batching, the agent has also internalized a confidence policy for action batching.

\subsection{Inference efficiency} \label{subsub: inference-efficiency}
To quantify the average latency reduction, we benchmark \ourmodel{} against a non-optimized baseline that processes every instance via a stepwise TableQA pipeline. For generalisability, both settings test on same modalities, including fixed and our proposed action-conditioned multimodal selection policy. Detailed evaluation protocol is in Appendix~\ref{sec:efficiency-benchmark-protocol}. The results presented in Figure~\ref{fig:main-ablation}(e) demonstrate substantial efficiency gains. Specifically, \ourmodel{} facilitates an average reduction of 33.35\% in latency compared to unoptimized baseline, decreasing mean inference time from 41.55s to 27.29s per sample, and average turns needed for reasoning from 3.61 turns to 2.24. While we record comparable computes due to additional state estimation need per turn, the TPS has significantly improved due to drop in average turns needed and inference latency. These improvements are direct consequence of \ourmodel{}, which reduces the heavy communication overhead of multi-turn conversation. 

\section{Summary and Future Works}
We present \ourmodel{} as a training-free framework for multi-turn table reasoning that optimizes reasoning action via representation and estimation. We first propose an action-conditional modality selection to improve table understanding, by routing each stepwise input to an optimal readout, i.e. text, image or multimodal, based on the action to be taken. To enhance stepwise reasoning efficiency, we incorporate state estimation for trajectory optimization, while proposing table metadata as a low-dimensional projection to preserve reliable state estimation of complex data. Our experiments across seven diverse benchmarks show that \ourmodel{} consistently achieves strong performance under a training-free setup, validating that its dynamic-modality and metadata-guided reasoning is an efficient paradigm for complex table reasoning tasks. Looking ahead, we position the evolution of \ourmodel{} to incorporate multimodality into chain-of-thoughts, as indicated by recent works on effectiveness of non-text reasoning~\citep{xu2026visual}, and to establish a new standard for reliable and efficient table reasoning.  
\section*{Ethics Statement}
This work studies how to improve table reasoning for language models in interactive multi-turn conversation environments. Our experiments exclusively use pubilcly available benchmark and packages, and do not involve any non-public data or personally identifiable information.

\section*{LLM Usage}
In this work, we utilize LLM for code debugging, codebase restructuring and writing polishing. Specifically, we also leverage the coding ability of LLM to generate structured figures (Figure~\ref{fig:atten-illus} and~\ref{fig:main-ablation}). We have prepared a repository restructured by LLM that contains 20 questions for all used datasets in this work for sample reproducibility. Details are in Appendix~\ref{sub: table-representation-attention}.

\bibliography{colm2026_conference}
\bibliographystyle{colm2026_conference}

\appendix
\section{Implementation Details} \label{sec: implementation-details}
\subsection{\ourmodel{} Algorithm}
In our experiment, we instantiate modality $m_{0}$ as image, which the MLLM agent would choose the modality $m$ in each step. 
\begin{algorithm}[H]
\caption{\ourmodel{}}
\label{alg:simple}
\begin{algorithmic}[1]
\small % Reduces font size slightly for better fit
\Require $q, df, \mathcal{M}, \mathcal{T}, T, m_{0}$ \Ensure $\hat{a}$
\State $env \gets df, m \gets m_{0}, fb_0 \gets \emptyset$ 
\For{$t = 1, \dots, T$}
\State $o_t \gets \begin{cases} \text{Ser}(env) & m = \text{txt} \\ \text{Rend}(env) \mathbin{\|} \text{cols} & m = \text{multi} \\ \text{Rend}(env) & m = \text{img} \end{cases}$
    \State $\mathcal{A}_t \gets \mathcal{M}(o_t, q, \mathcal{T}, fb_{t-1})$
    \For{each $a \in \mathcal{A}_t$}
        \If{$a = \text{f\_ans}$} \Return $a.ans$ \EndIf
        \If{$a = \text{f\_swi}$} $m \gets a.mode$ 
        \Else \ $(env, out) \gets \text{Exec}(a, env)$ \EndIf
    \EndFor
    \State $fb_t \gets (out, \text{Rew}(env), \text{Score}(a.pred, env))$
\EndFor
\end{algorithmic}
\end{algorithm}

\subsection{Inference Prompts} \label{sub: inference-prompt}
We follow standardized prompting on all testing datasets, leaving the agent to answer the query as-is. 
\subsubsection{Role and Default Prompt}
\vspace{5mm}
\begin{sysbox}
\medskip
You are a table reasoning agent. Use tools to transform the table and answer the question.\\
You may batch multiple simple actions together in one step to save time.

\medskip
Respond with JSON only using the schema:
\begin{verbatim}
{
  "reasoning": str,
  "predicted_metadata": {
      "rows": "fewer|same|more",// or exact integer
      "cols": [str] or null,    // expected column names, 
      "key_output": str or null // expected value in tool output
  },
  "actions": [{"tool_name": str, "tool_args": object}, ...]
}
\end{verbatim}

\medskip
\textbf{Action batching rules:}
\begin{itemize}[nosep, leftmargin=1.5em]
  \item Group low-complexity operations in one step (e.g.\ filter $+$ sort, select $+$ rename).
  \item Use a single action when the result of one step determines the next
        (e.g.\ check row count after a filter).
  \item To answer, include \field{\{"tool\_name": "f\_final\_answer", "tool\_args": \{"answer": "..."\}\}}
        as the \emph{last} action.
  \item To revert to the original table, use \field{f\_retrieve\_original\_df}.
\end{itemize}
\end{sysbox}

\subsubsection{Table Observation}
\vspace{5mm}
\begin{adaptbox}
Table observation is prepended to the message modality-dependent
\medskip
\begin{tabular}{lp{9.5cm}}
\toprule
\textbf{Mode} & \textbf{What is attached} \\
\midrule
\texttt{image}      & JPEG render of \texttt{env[df]} (200\,dpi, up to 30 rows $\times$ 5 cols)
                      prepended as \texttt{<image>} token \\
\texttt{multimodal} & Same JPEG \emph{plus} column names appended as text \\
\texttt{text}       & Markdown table (via \texttt{to\_markdown}); no image \\
\bottomrule
\end{tabular}
\end{adaptbox}

\bigskip

\subsubsection{Step Prompt}
\vspace{5mm}
\begin{dynbox}
\medskip
\placeholder{step budget note $b_t$}
\quad — one of:
\begin{itemize}[nosep, leftmargin=1.5em]
  \item \texttt{[URGENT] This is your last step ($t$/$T$). You MUST call f\_final\_answer now.}
  \item \texttt{[NOTE] $k$ steps remaining ($t$/$T$). Start converging toward f\_final\_answer.}
  \item \texttt{Step $t$ of $T$.}
\end{itemize}

\medskip
\field{Question:} \placeholder{$q$}\\[2pt]
\field{Table shape:} \placeholder{$(r, c)$}\\[2pt]
\field{Columns:} \placeholder{$[$col$_1$, col$_2$, \ldots$]$}\\[2pt]
\field{Preview rows (JSON):} \placeholder{first 15 rows of \texttt{env[df]} as JSON}\\[2pt]
\field{Last step output:} \placeholder{truncated to 400 chars; \texttt{N/A} at $t=1$}\\[2pt]
\field{Last error:} \placeholder{\texttt{N/A} if none}

\medskip
\textbf{Cross-step action ledger} \quad\textcolor{gray}{\small(drift guard)}\\
\placeholder{Prior actions this episode:}\\
\quad\texttt{step$k$: tool\_name(args)}
  \quad\textit{$\leftarrow$ REPEATED $n\times$, avoid calling again} \quad (if $n \ge 2$)\\
\textcolor{borderred}{\small[DRIFT WARNING] tool+args combos called 2$+$ times: \ldots
  You MUST try a different approach or call f\_final\_answer.}
  \quad (if any repeats)

\medskip
\textbf{Tools:}\\
\placeholder{formatted tool catalogue $\mathcal{S}$ — name, description, parameter list}
\end{dynbox}

\medskip

\subsection{Datasets and Metrics}\label{app:datasets-and-metrics}
Our experimental evaluation is conducted on seven diverse benchmarks across TQA and TFV. We follow the established protocol of previous works~\citep{zheng-etal-2024-multimodal,wang2026hippoenhancingtableunderstanding,xing2026tabledart} for evaluation metrics, using Accuracy for most tasks and BLEU score for the generative FeTaQA benchmark. Table~\ref{tab:dataset-stats} provides a detailed breakdown of these datasets, including the number of instances used for our experiment.%, \textcolor{red}{where the training and validation set is for RL training. }

\begin{table}[ht]
\centering
\caption{Dataset statistics by task type.}
\label{tab:dataset-stats}
\renewcommand{\arraystretch}{1.15}
\setlength{\tabcolsep}{10pt}
\begin{tabular}{llccc}
\toprule
\textbf{Task Type} & \textbf{Dataset} & \textbf{Train} & \textbf{Validation} & \textbf{Test} \\
\midrule
\multirow{5}{*}{\makecell[l]{Table Question\\Answering (TQA)}} 
& TABMWP & 500 & 200 & 1,000 \\
& WTQ    & 500 & 200 & 1,000 \\
& HiTab  & 500 & 200 & 1,000 \\
& TAT-QA & --- & --- & 772 \\
& FeTaQA & 500 & 200 & 1,000 \\
\midrule
\multirow{2}{*}{\makecell[l]{Table Fact\\Verification (TFV)}} 
& TabFact  & 500 & 200 & 1,000 \\
& InfoTabs & 500 & 200 & 1,000 \\
\midrule
\textbf{Total} &  & \textbf{3,000} & \textbf{1,400} & \textbf{6,772} \\
\bottomrule
\end{tabular}
\end{table}

\paragraph{Ablation datasets.} Aside from WikiTQ which is a standard tableQA benchmark, we particularly conduct our ablation studies on MMQA and MMTU, which are two datasets released at 2025, so as to simulate a scenario where the backbone LLM has not seen the datasets during pre-training. MMQA is a large-scale benchmark for evaluating LLMs on multi-table and multi-hop question answering. The benchmark includes a total of 3,312 relational tables across 138 domains, where each instance consists of two or three interlinked tables. The dataset features 5,000 multi-table samples, annotated with natural language questions, SQL queries, gold answers, and explicit primary/foreign key relations. To ensure annotation quality, foreign and primary keys were labeled by human experts with inter-annotator agreement exceeding 80\%. MMQA questions span four main categories, including numerical, list, count, and select, with an average length of 77–85 tokens, reflecting their compositional complexity. MMTU is a large-scale benchmark with around 28K questions across 25 real-world table tasks. These tasks are drawn from decades' worth of computer science research on tabular data, with a focus on complex table tasks faced by professional users, where even frontier reasoning models like OpenAI GPT-5 and DeepSeek R1 score only around 69\% and 57\% respectively, suggesting significant room for improvement. For MMTU, we specifically extract the question subgroups Table-QA, Table-Fact-Verification and Table-Lookup to replicate the same table reasoning tasks only on datasets unreleased by the time when the base models were released. While the two datasets are released recently so that most baseline works have yet to evaluate on, we conduct a detailed ablation study to report the base performance of existing baselines on these two state-of-the-art TableQA datasets. 

\subsection{Comparative Models and Rationale}
\label{app:comparative-models}

We compare \ourmodel{} against a comprehensive set of baselines, selected to validate our core contributions across multiple axes.

\paragraph{Constituent models.}
To demonstrate that \ourmodel{}'s performance arises from its dynamic framework rather than any single component, we evaluate its constituent models as standalone baselines. These include: Qwen2.5-VL-7B~\citep{bai2025qwen25vltechnicalreport} and Qwen3-VL-8B~\citep{bai2025qwen3vltechnicalreport}. Evaluating the VLM in isolation establishes direct performance baselines. 

\paragraph{Multimodal baselines.}
To compare the advantage of action-conditioned multimodal selection over static, one-modality-fits-one-attempt approaches, we compare against two multimodal LLM baselines and recent work on dynamic adaptive routing TableDART~\citep{xing2026tabledart}. The first MLLM is HIPPO~\citep{wang2026hippoenhancingtableunderstanding} which jointly processes both text and image representations for all inputs. The second is Gemini 2.0 Flash~\citep{google2025gemini25} which serves as an additional MLLM baseline in regards to its constituent support on TableDART fusion path. Lastly, we compare against the state-of-the-art TableDART which routes modality in each question attempt with a trained model for isolate comparison between advantage of selecting modality conditioning on each action or on each separate question attempt. 

\paragraph{Trajectory optimizing baselines.}
To compare the advantage of our proposed metadata-guided  trajectory optimization over existing trajectory optimziation methods, we compare against two key baselines which optimize trajectory via process reward model~\citep{zou2026tattoo} and stepwise table pruning~\citep{guo2026rethinkingtablepruningtableqa}. The first work trains a process reward model from Qwen3-8B~\citep{yang2025qwen3technicalreport} to provide stepwise reward evaluation, while the second work trains a pruner from derived gold pruning trajectory. We initiate these comparisons to evaluate the effectiveness of projecting high-dimensional structured data to low-dimensional metadata for reward modeling. 

\paragraph{Broader competitive landscape.}
To position \ourmodel{} within the broader literature, we benchmark it against an extensive suite of single-modality baselines in the 7--8B parameter range, including table-as-text baselines under standard LLMs such as Llama-2-7B~\citep{touvron2023llama2openfoundation}, Llama3-Instruct-8B~\citep{grattafiori2024llama3herdmodels}, TableLlama-7B~\citep{zhang-etal-2024-tablellama}. We also include various table-as-image models aside from our base constituent models with Table-LLaVA-7B~\citep{zheng-etal-2024-multimodal}, SynTab-LLaVA-7B~\citep{11093154}, MiniCPM-V-2.6-8B~\citep{minicpm}. Additional training-free TableQA agent baselines are also included, containing DATER~\citep{10.1145/3539618.3591708}, ReAcTable~\citep{10.14778/3659437.3659452}, Mix-SC~\citep{liu-etal-2024-rethinking}, TIDE~\citep{yang2025triples} and CIT-DP~\citep{yang2025causality}

\paragraph{Result sources and reporting.}
For a comprehensive and fair comparison, we report several baseline results directly from prior work. Specifically, we adopt Llama2-7B, Llama3-Instruct-8B, TableLlama-7B, {TableLLaVA-7B}, {MiniCPM-V-2.6-8B}, and \textsc{HIPPO} are adopted from \citep{wang2026hippoenhancingtableunderstanding}. Results for dynamic adaptive routing TableDART are taken from its proposal paper~\citep{xing2026tabledart}, while results for trained trajectory optimizing agents are taken from the proposed TATTOO PRM paper~\citep{zou2026tattoo} and TabTrim pruner paper~\citep{guo2026rethinkingtablepruningtableqa}. Lastly, the training-free TableQA agents are taken from recent published paper~\citep{yang2025causality}. All remaining results not covered above are generated from our own experimental runs.

\subsection{Hyperparameter Configuration}
Table~\ref{tab:all_hyperparameters} summarizes the complete hyperparameter settings for experiments. These hyperparameters are selected through preliminary experiments to balance inference time and efficiency. 
\begin{table*}[ht]
\centering
\caption{Summary of image rendering, model generation, and experiment hyperparameters.}
\label{tab:all_hyperparameters}
\renewcommand{\arraystretch}{1.12}
\setlength{\tabcolsep}{8pt}
\resizebox{0.95\textwidth}{!}{%
\begin{tabular}{lll}
\toprule
\textbf{Parameter} & \textbf{Value / Default} & \textbf{Description} \\
\midrule
\multicolumn{3}{c}{\textbf{Image Rendering Hyperparameters}} \\
\midrule
Max rows displayed & 30 & Truncate overly wide tables\\
Max cols displayed & 10 & Truncate overly long tables\\
Font size & 8pt & Standardize patch size for each word\\
Row scale & 1.2$\times$ & Standardize patch size for each row\\
DPI & 200 & Standardize resolution\\
Format & \texttt{.jpg} & Standardize image format\\
\midrule
\multicolumn{3}{c}{\textbf{Model / Generation Hyperparameters}} \\
\midrule
{Max tokens}  & 8192 & Max output tokens per call \\
{Temperature} & 0.7 & Sampling temperature (Mandatory 1 for GPT-5 models)\\
{Seed}        & 42 & Random seed (optional) \\
\midrule
\multicolumn{3}{c}{\textbf{Experiment / Agentic Loop Hyperparameters}} \\
\midrule
{Max steps} & 12 & Max tool-use steps per question \\
{Runtime retries} & 2 & Retries per step for JSON parse errors \\
{Attempts per question} & 1 & Independent attempts per question \\
{Seed}        & 42 & Random seed (optional) \\
\bottomrule
\end{tabular}}
\end{table*}

\subsection{Computational Configuration}
Inference is conducted on a 2 NVIDIA RTX4090 GPU with FP16 precision, as optimized by Ollama~\citep{ollama2025}. %\textcolor{red}{RL training is done on 8 NVIDIA H200 120GB GPU with bfloat16 precision to improve training efficiency. The complete training process requires approximately 26 hours to complete one epoch. Training converges rapidly within the epoch, with evaluation done on the held-out validation set. Checkpoint with the highest validation accuracy is used for inference. }

\subsection{Efficiency Benchmark Protocol}
\label{sec:efficiency-benchmark-protocol}
This section provides a detailed account of the protocol used for the efficiency analysis presented in Section~\ref{subsub: inference-efficiency}. We include data sampling methodology and precise definitions for reported metrics to ensure full reproducibility, as reference to protocol in prior work~\citep{xing2026tabledart}.

\subsubsection{Benchmark Setup and Data Sampling}
To create a representative testbed, we constructed an evaluation set via stratified random sampling from the three benchmark test sets. We randomly sampled a balanced set of 200 instances from each dataset, resulting in a comprehensive benchmark suite of 600 unique samples. To ensure statistical stability, the entire measurement process was repeated three times with different random seeds, and all reported metrics are the average across these independent runs. All benchmarks were executed on a single NVIDIA RTX4090 GPU.

\subsubsection{Metric Definitions}
We use the following three primary metrics to report efficiency:

\begin{itemize}
    \item \textbf{Conversation Turns:} The total conversation turns to complete the stepwise reasoning, calculated as number of actions. This helps us identify the effectiveness of action compression in combining multiple actions into one. Lower values are better.
    \item \textbf{Latency (s):} Total time in seconds to process a single sample. This is our primary metric to evaluate end-to-end inference speed. Lower values are better.
    \item \textbf{Tokens per Second (TPS):} A measure of throughput calculated by dividing number of generated output tokens, including all reasoning tokens, by the total latency. The token count is collected during generation. Higher values are better.
\end{itemize}

\section{Supporting Information of Section \ref{sub: ablation} Ablation Studies} \label{sec: main-results-detailed}
\subsection{Details of Section \ref{sub: ablation} Ablation Studies}
\begin{figure}[t]
    \centering
    \includegraphics[width=0.95\linewidth]{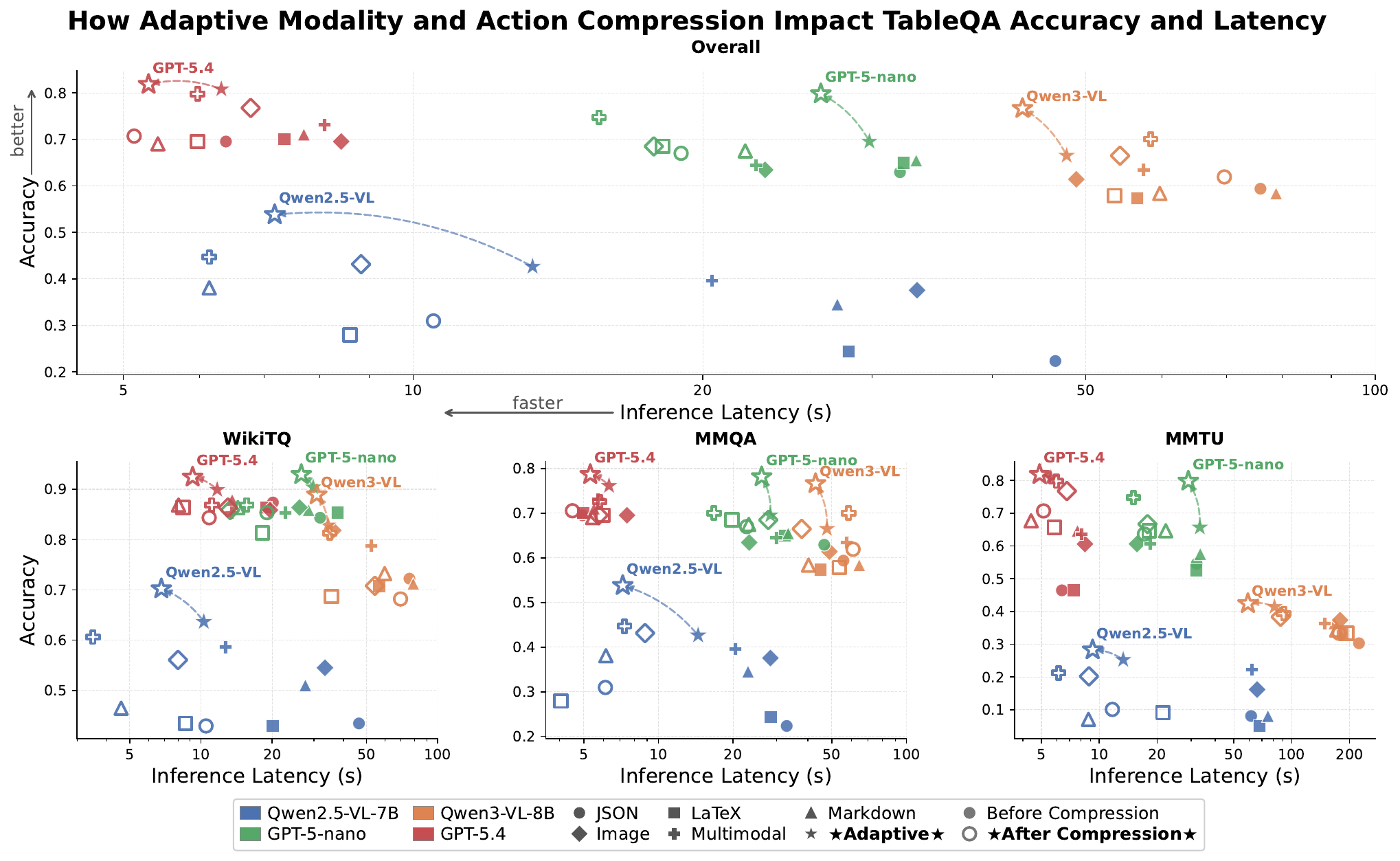}
    \caption{Overall performance of all models in the three datasets.}
    \label{fig:main-result-overall}
\end{figure}

This section includes detailed numerical figures for Main Results (Figure~\ref{fig:main-ablation}). Specifically, Figure~\ref{fig:main-result-overall} illustrates the specific performance on each dataset, where similar performance patterns are observed across the three datasets, indicating the dataset-agnostic improvement by \ourmodel{}. Most notably, we observe that the individual improvement brought by utilizing the action-conditioned multimodal selection policy shows a more significant improvement in three small models compared to the large proprietary GPT-5.4. This suggests that larger model is naturally more capable of understanding text serialized tabular data and attending to more distant tokens, offsetting inherent advantage of the action-conditioned multimodal selection policy. Nevertheless, under a resource-constrained budget, we observe a consistent accuracy improvement in having action-conditioned multimodal to represent tabular state. 

\begin{table}[H]
\centering
\caption{Detailed accuracy of Figure \ref{fig:main-ablation}(a) and Figure \ref{fig:main-result-overall}}
\label{tab:main_results}

\setlength{\tabcolsep}{3pt}
\renewcommand{\arraystretch}{1.15}

\resizebox{0.9\textwidth}{!}{%
\begin{tabular}{l|c|c|c|c|c|c|c|c|c|c|c|c|c}
\toprule
\multicolumn{14}{c}{\Large{\textbf{Accuracy}}}\\
\midrule
Modality & \texttt{json} & \texttt{latex} & \texttt{Markdown} & Image & Multimodal
& \hilight{Adaptive}
& \texttt{json} & \texttt{latex} & \texttt{Markdown} & Image & Image & \hilight{Multimodal}
&  \hilite{Adaptive}
\\
Optimization & \cross & \cross & \cross & \cross & \cross & \hilight{\cross} &\tick&\tick&\tick&\tick (Metadata)&\tick (Table)&\hilight{\tick}&\hilite{\tick}\\
\midrule
\multicolumn{14}{c}{WTQ}\\
\midrule
 Qwen2.5-VL-7B & 0.4343 & 0.4293 & 0.5101 & 0.5450 & 0.5859 & \hilight{0.6364} & 0.4293 & 0.4343 & 0.4646 & 0.5606 & 0.4814 &\hilight{0.6061} & \hilite{\textbf{0.7020}}\\
Qwen3-VL-8B & 0.7222& 0.7071 & 0.7121 & 0.8182 & 0.7929 & \hilight{0.8333} & 0.7071 & 0.6869 & 0.7323 & 0.7980 & 0.7525 & \hilight{0.8131} & \hilite{\textbf{0.8889}}\\
 GPT-5-nano & 0.8434 & 0.8535 & 0.8586 & 0.8636 & 0.8535 & \hilight{0.9040} & 0.8535 & 0.8131 & 0.8636 & 0.8586 & 0.8131 & \hilight{0.8687} & \hilite{\textbf{0.9293}}\\
 GPT-5.4 & 0.8737 & 0.8636 & 0.8788 & 0.8586 & 0.8535 & \hilight{0.8990} & 0.8434 & 0.8636 & 0.8687 & 0.8636 & 0.8384 & \hilight{0.8687} & \hilite{\textbf{0.9242}}\\

\midrule
\multicolumn{14}{c}{MMQA}\\
\midrule
 Qwen2.5-VL-7B & 0.2234 & 0.2437 & 0.3452 & 0.3756 & 0.3959 & \hilight{0.4264} & 0.3096 & 0.2792 & 0.3807 & 0.4315 & 0.3056 & \hilight{0.4467} & \hilite{\textbf{0.5381}}\\
Qwen3-VL-8B & 0.5939 & 0.5736 & 0.5838 & 0.6142 & 0.6345 & \hilight{0.6650} & 0.6193 & 0.5787 & 0.5838 & 0.6650 & 0.5431 & \hilight{0.7005} & \hilite{\textbf{0.7665}} \\

 GPT-5-nano & 0.6294 & 0.6497 & 0.6548 & 0.6345 & 0.6447 & \hilight{0.6954} & 0.6701 & 0.6853 & 0.6751 & 0.6850 & 0.6294 & \hilight{0.7005} & \hilite{\textbf{0.7817}}\\
GPT-5.4 & 0.6954 & 0.7005 & 0.7107 & 0.6954 & 0.7310 & \hilight{0.7614} & 0.7056 & 0.6954 & 0.6904 & 0.6954 & 0.6244 & \hilight{0.7259} & \hilite{\textbf{0.7868}}\\

\midrule 
\multicolumn{14}{c}{MMTU}\\
\midrule
Qwen2.5-VL-7B & 0.0808 & 0.0505 & 0.0808 & 0.1616 & 0.2222 & \hilight{0.2525} & 0.1010 & 0.0909 & 0.0707 & 0.2020 & 0.1053 &\hilight{0.2121} & \hilite{\textbf{0.2828}}\\
Qwen3-VL-8B & 0.3030 & 0.3333 & 0.3636 & 0.3737  & 0.3636 & \hilight{0.4141} & 0.3333 & 0.3333 & 0.3434 & 0.3838 & 0.3030 & \hilight{0.3939} & \hilite{\textbf{0.4242}} \\
 GPT-5-nano & 0.5455 & 0.5253 & 0.5758 & 0.6061 & 0.6061 & \hilight{0.6566} & 0.6364 & 0.6465 & 0.6465 & 0.6667 & 0.4646 & \hilight{0.7475} & \hilite{\textbf{0.7980}}\\
GPT-5.4 & 0.4646 & 0.4646 & 0.6465 & 0.6061 & 0.6364 & \hilight{0.8081} & 0.7071 & 0.6566 & 0.6768 & 0.7677 & 0.5253 & \hilight{0.7980} & \hilite{\textbf{0.8182}}\\

\bottomrule
\end{tabular}}
\vspace{-\baselineskip}
\end{table}

\begin{table}[H]
\centering
\caption{Detailed standard deviation for Figure \ref{fig:main-ablation}(a) and Figure \ref{fig:main-result-overall}}
\label{tab:main_results_sd}

\setlength{\tabcolsep}{3pt}
\renewcommand{\arraystretch}{1.15}

\resizebox{0.9\textwidth}{!}{%
\begin{tabular}{l|c|c|c|c|c|c|c|c|c|c|c|c|c}
\toprule
\multicolumn{14}{c}{\Large{\textbf{Standard Deviation}}}\\
\midrule
Modality & \texttt{json} & \texttt{latex} & \texttt{Markdown} & Image & Multimodal
& \hilight{Adaptive}
& \texttt{json} & \texttt{latex} & \texttt{Markdown} & Image & Image & \hilight{Multimodal}
&  \hilite{Adaptive}
\\
Optimization & \cross & \cross & \cross & \cross & \cross & \hilight{\cross} &\tick&\tick&\tick&\tick (Metadata)&\tick (Table)&\hilight{\tick}&\hilite{\tick}\\
\midrule
\multicolumn{14}{c}{WTQ}\\
\midrule
 Qwen2.5-VL-7B & 0.0352 & 0.0352 & 0.0355 & 0.0354 & 0.0350 & \hilight{0.0342} & 0.0352 & 0.0352 & 0.0354 & 0.0353 & 0.0355 & \hilight{0.0347} & \hilite{\textbf{0.0325}} \\
Qwen3-VL-8B & 0.0318 & 0.0323 & 0.0322 & 0.0274 & 0.0288 & \hilight{0.0265} & 0.0323 & 0.0330 & 0.0315 & 0.0285 & 0.0307 & \hilight{0.0277} & \hilite{\textbf{0.0223}}\\
 GPT-5-nano & 0.0258 & 0.0251 & 0.0248 & 0.0244 & 0.0251 & \hilight{0.0209} & 0.0251 & 0.0277 & 0.0244 & 0.0248 & 0.0277 & \hilight{0.0240} & \hilite{\textbf{0.0182}}\\
 GPT-5.4 & 0.0236 & 0.0244 & 0.0232 & 0.0248 & 0.0251 & \hilight{0.0214} & 0.0258 & 0.0244 & 0.0240 & 0.0244 & 0.0262 & \hilight{0.0240} & \hilite{\textbf{0.0188}}\\

\midrule
\multicolumn{14}{c}{MMQA}\\
\midrule
 Qwen2.5-VL-7B & \textbf{0.0299} & 0.0308 & 0.0339 & 0.0345 & 0.0348 & \hilight{0.0352} & 0.0329 & 0.0320 & 0.0346 & 0.0353 & 0.0328 & \hilight{0.0354} & \hilite{0.0356}\\
Qwen3-VL-8B & 0.0350 & 0.0352 & 0.0351 & 0.0347 & 0.0343 & \hilight{0.0336} & 0.0346 & 0.0352 & 0.0351 & 0.0336 & 0.0355 & \hilight{0.0326} & \hilite{\textbf{0.0301}} \\
 GPT-5-nano & 0.0344 & 0.0340 & 0.0339 & 0.0343 & 0.0341 & \hilight{0.0328} & 0.0335 & 0.0331 & 0.0334 & 0.0331 & 0.0344 & \hilight{0.0326} & \hilite{\textbf{0.0294}}\\
GPT-5.4 & 0.0328 & 0.0326 & 0.0323 & 0.0328 & 0.0316 & \hilight{0.0304} & 0.0325 & 0.0328 & 0.0329 & 0.0328 & 0.0345 & \hilight{0.0318} & \hilite{\textbf{0.0292}}\\

\midrule 
\multicolumn{14}{c}{MMTU}\\
\midrule
Qwen2.5-VL-7B & 0.0193 & \textbf{0.0155} & 0.0193 & 0.0260 & 0.0294 & \hilight{0.0307} & 0.0213 & 0.0203 & 0.0181 & 0.0284 & {0.0217} & \hilight{0.0289} & \hilite{0.0318}\\
Qwen3-VL-8B & \textbf{0.0327} & 0.0336 & 0.0343 & 0.0345 & 0.0343 & \hilight{0.0351} & 0.0336 & 0.0336 & 0.0338 & 0.0346 & \textbf{0.0327} & \hilight{0.0348} & \hilite{0.0352}\\
 GPT-5-nano & 0.0352 & 0.0353 & 0.0349 & 0.0346 & 0.0346 & \hilight{0.0336} & 0.0340 & 0.0338 & 0.0338 & 0.0333 & 0.0353 & \hilight{0.0307} & \hilite{\textbf{0.0284}}\\
GPT-5.4 & 0.0353 & 0.0353 & 0.0338 & 0.0346 & 0.0340 & \hilight{0.0278} & 0.0322 & 0.0336 & 0.0331 & 0.0299 & 0.0353 & \hilight{0.0284} & \hilite{\textbf{0.0273}}\\

\bottomrule
\end{tabular}}
\vspace{-\baselineskip}
\end{table}

\begin{table}[H]
\centering
\caption{Detailed time latency of Figure \ref{fig:main-ablation}(a) and Figure \ref{fig:main-result-overall}}
\label{tab:main_inference}

\setlength{\tabcolsep}{3pt}
\renewcommand{\arraystretch}{1.15}

\resizebox{0.9\textwidth}{!}{%
\begin{tabular}{l|c|c|c|c|c|c|c|c|c|c|c|c|c}
\toprule
\multicolumn{14}{c}{\Large{\textbf{Inference Latency (in seconds)}}}\\
\midrule
Modality & \texttt{json} & \texttt{latex} & \texttt{Markdown} & Image & Multimodal
& \hilight{Adaptive}
& \texttt{json} & \texttt{latex} & \texttt{Markdown} & Image & Image & \hilight{Multimodal}
&  \hilite{Adaptive}
\\
Optimization & \cross & \cross & \cross & \cross & \cross & \hilight{\cross} &\tick&\tick&\tick&\tick (Metadata)&\tick (Table)&\hilight{\tick}&\hilite{\tick}\\
\midrule
\multicolumn{14}{c}{WTQ}\\
\midrule
 Qwen2.5-VL-7B & 46.47 & 20.05 & 27.62 & 33.35 & 12.73 & \hilight{10.28} & 10.50 & 8.58 & 4.59 & 7.95 & 12.68 & \hilight{\textbf{3.50}} & \hilite{6.75}\\
Qwen3-VL-8B & 75.96 & 56.56 & 78.85 & 36.22 & 52.48 &\hilight{34.44} & 69.67 & 35.59 & 59.70 & 54.28 & 40.89 & \hilight{35.00} & \hilite{\textbf{30.83}}\\
 GPT-5-nano & 31.86 & 37.71 & 28.48 & 26.07 & 22.73 & \hilight{29.80} & 18.99 & 18.17 & 14.31 & \textbf{13.27} & 27.34 & \hilight{15.61} & \hilite{26.53}\\
 GPT-5.4 & 20.13 & 18.82 & 13.55 & 19.58 & 13.47 & \hilight{11.71} & 10.83 & 8.40 & \textbf{8.04} & 13.00 & 8.09 & \hilight{11.12} & \hilite{9.22}\\

\midrule
\multicolumn{14}{c}{MMQA}\\
\midrule
 Qwen2.5-VL-7B & 32.88 & 28.36 & 22.97 & 28.27 & 20.44 & \hilight{14.45} & 6.11 & \textbf{4.04} & 6.14 & 8.83 & 6.83 & \hilight{7.30} & \hilite{7.18}\\
Qwen3-VL-8B & 55.68 & 45.00 & 64.48 & 48.89 & 57.44 & \hilight{47.77} & 60.89 & 53.58 & 40.28 & 37.79 & 36.01 & \hilight{58.42} & \hilite{\textbf{42.98}}\\
 GPT-5-nano & 46.59 & 32.34 & 33.33 & 23.22 & 29.87 & \hilight{28.24} & 22.68 & 19.77 & 23.12 & 27.72 & 20.98 & \hilight{\textbf{16.75}} & \hilite{26.06}\\
GPT-5.4 & 4.94 & 4.97 & 5.50 & 7.47 & 5.72 & \hilight{6.32} & \textbf{4.50} & 5.97 & 5.43 & 5.74 & 5.66 & \hilight{5.79} & \hilite{5.31}\\

\midrule 
\multicolumn{14}{c}{MMTU}\\
\midrule

Qwen2.5-VL-7B & 61.53 & 68.03 & 75.26 & 66.10 & 62.18 & \hilight{13.31} & 11.69 & 21.40 & 8.78 & 8.83 & 8.29 & \hilight{\textbf{6.14}} & \hilite{9.23}\\
Qwen3-VL-8B & 223.76 &  183.87 & 166.59 & 178.51 & 148.91 & \hilight{81.37} & 176.81 & 193.02 & 171.19 & 87.65 & 53.17 & \hilight{90.83} & \hilite{\textbf{59.22}}\\
 GPT-5-nano & 32.06 & 31.86 & 33.47 & 15.69 & 18.36 & \hilight{33.30} & 17.16 & 18.15 & 22.14 & 17.79 & 39.01 & \hilight{\textbf{15.05}} & \hilite{29.02}\\
GPT-5.4 & 6.39 & 7.35 & 7.70 & 8.42 & 8.09 & \hilight{5.60} & 5.13 & 5.84 & \textbf{4.42} & 6.78 & 9.61 & \hilight{5.97} & \hilite{4.89}\\

\bottomrule
\end{tabular}}
\vspace{-\baselineskip}
\end{table}

\subsection{Supporting ROUGE and BLEU evaluation for Section \ref{sub: ablation}}\label{sub: main-rouge}
In our additional evaluation with ROUGE~\citep{rouge} and BLEU~\citep{10.3115/1073083.1073135}, we observe that ROUGE score indicates similar performance pattern as the accuracy-based evaluation, with a much less consistent result on the BLEU evaluation metrics. This suggests that lexical overlapping metrics may not be well suited for evaluating TableQA correctness, as they fail to measure the semantic equivalence of the generated answer, especially in scenarios of paraphrasing, formatting variation or abbreviations. As ROUGE shows to be more consistent with the accuracy results, we attribute this to BLEU's design for machine translation and emphasis of exact $n$-gram precision. This renders BLEU harsh on cases with short ground truth, which happen to be the case given existing TableQA datasets. 
\begin{table}[t]
\centering
\caption{ROUGE~\citep{rouge} and BLEU~\citep{10.3115/1073083.1073135} score of Main Result Figure~\ref{fig:main-ablation}(a)}
\label{tab:main_rouge}

\setlength{\tabcolsep}{3pt}
\renewcommand{\arraystretch}{1.15}

\resizebox{0.95\textwidth}{!}{%
\begin{tabular}{l|c|c|c|c|c|c|c|c|c|c|c|c|c}
\toprule
\multicolumn{14}{c}{\Large{\textbf{ROUGE / \small{BLEU}}}}\\
\midrule
Modality & \texttt{json} & \texttt{latex} & \texttt{Markdown} & Image & Multimodal
& \hilight{Adaptive}
& \texttt{json} & \texttt{latex} & \texttt{Markdown} & Image & Image & \hilight{Multimodal}
&  \hilite{Adaptive}
\\
Optimization & \cross & \cross & \cross & \cross & \cross & \hilight{\cross} &\tick&\tick&\tick&\tick (Metadata)&\tick (Table)&\hilight{\tick}&\hilite{\tick}\\
\midrule
\multicolumn{14}{c}{WTQ}\\
\midrule
 Qwen2.5-VL-7B & 0.608 / \small{0.029} & 0.618 / \small \textbf{0.039} & 0.580 / \small 0.035 & 0.599 / \small 0.037 & 0.640 / \small 0.024 & \hilight{0.582 / \small 0.026} & 0.546 / \small 0.009 & 0.496 / \small 0.024 & 0.512 / \small 0.016 & 0.573 / \small 0.024 & 0.467 / \small 0.000 & \hilight{0.565 / \small 0.029} & \hilite{\textbf{0.642} / \small 0.038}\\
Qwen3-VL-8B & 0.709 / \small 0.038 & 0.713 / \small 0.030 & 0.715 / \small 0.028 & 0.769 / \small 0.034 & 0.768 / \small 0.034 & \hilight{0.810 / \small \textbf{0.040}} & 0.717 / \small 0.036 & 0.722 / \small 0.031 & 0.753 / \small 0.034 & 0.774 / \small 0.026 & 0.780 / \small 0.031 & \hilight{0.775 / \small 0.035} & \hilite{\textbf{0.852} / \small 0.035}\\
 GPT-5-nano & 0.745 / \small \textbf{0.045} & 0.742 / \small 0.033 & 0.758 / \small 0.037 & 0.764 / \small 0.031 & 0.746 / \small 0.025 & \hilight{0.785 / \small 0.040} & 0.772 / \small 0.035 & 0.732 / \small 0.040 & 0.738 / \small 0.037 & 0.734 / \small 0.038 & 0.701 / \small 0.031 & \hilight{0.744 / \small 0.040} & \hilite{\textbf{0.793} / \small 0.039}\\
 GPT-5.4 & 0.764  / \small 0.031 & 0.746 / \small \textbf{0.034} & 0.760 / \small 0.033 & 0.717 / \small 0.033 & 0.711 / \small 0.026 & \hilight{\textbf{0.794} / \small 0.032} & 0.704 / \small \textbf{0.034} & 0.705 / \small 0.029 & 0.736 / \small 0.031 & 0.690 / \small 0.032 & 0.727 / \small 0.024 & \hilight{0.679 / \small 0.030} & \hilite{0.784 / \small 0.031}\\

\midrule
\multicolumn{14}{c}{MMQA}\\
\midrule
 Qwen2.5-VL-7B & 0.320 / \small 0.030 & 0.398 / \small 0.049 & 0.404 / \small 0.034 & 0.393 / \small 0.024 & 0.407 / \small 0.027 & \hilight{0.436 / \small 0.030} & 0.387 / \small 0.039 & 0.353 / \small 0.035 & 0.396 / \small 0.033 & 0.385 / \small 0.032 & 0.327 / 0.004 & \hilight{0.402 / \small 0.040} & \hilite{\textbf{0.477} / \small \textbf{0.045}}\\
Qwen3-VL-8B & 0.590 / \small 0.054 & 0.566 / \small 0.038 & 0.556 / \small 0.046 & 0.584 / \small 0.051 & 0.579 / \small 0.043 & \hilight{0.615 / \small 0.047 }& 0.585 / \small 0.046  & 0.568 / \small 0.047 & 0.588 / \small 0.050 & 0.630 / \small \textbf{0.051} & 0.564 / \small 0.044 & \hilight{0.638 / \small 0.048} & \hilite{\textbf{0.698} / \small 0.049} \\

 GPT-5-nano & 0.572 / \small \textbf{0.057} & 0.590 / \small 0.055 & 0.590 / \small 0.054 & 0.588 / \small 0.048 & 0.605 / \small 0.054 & \hilight{0.636 / \small 0.050} & 0.594 / \small 0.050 & 0.599 / \small 0.054 & 0.573 / \small 0.048 & 0.580 / \small 0.045 & 0.586 / \small 0.036 & \hilight{0.596 / \small 0.048} & \hilite{\textbf{0.669} / \small 0.056}\\
GPT-5.4 & 0.590 / \small \textbf{0.054} & 0.592 / \small 0.053 & 0.585 / \small 0.049 & 0.589 / \small 0.043 & 0.594 / \small 0.046 & \hilight{\textbf{0.626} / \small 0.048} & 0.553 / \small 0.039 & 0.540 / \small 0.031 & 0.572 / \small 0.041 & 0.558 / \small 0.046 & 0.562 / \small 0.045 & \hilight{0.574 / \small 0.049} & \hilite{0.579 / \small 0.047}\\

\midrule 
\multicolumn{14}{c}{MMTU}\\
\midrule
Qwen2.5-VL-7B & 0.079 / \small 0.000 & 0.084 / \small 0.000 & 0.087 / \small 0.000 & 0.145 / \small 0.000 & 0.166 / \small 0.000 & \hilight{\textbf{0.218} / \small 0.000} & 0.121 / \small 0.000 & 0.101 / \small 0.000 & 0.067 / \small 0.000 & 0.132 / \small 0.000 & 0.082 / \small 0.000 & \hilight{0.177 / \small 0.000} & \hilite{0.206 / \small 0.000}\\
Qwen3-VL-8B & 0.258 / \small 0.000 & 0.296 / \small 0.000 & 0.321 / \small 0.000 & 0.309 / \small 0.000 & 0.315 / \small 0.000 & \hilight{0.341 / \small 0.000} & 0.283 / \small 0.000 & 0.288 / \small 0.000	 & 0.370 / \small 0.000 & 0.327 / \small 0.000 & 0.306 / \small 0.000 & \hilight{0.330 / \small 0.000} & \hilite{\textbf{0.355} / \small 0.000} \\
 GPT-5-nano & 0.276 / \small 0.000 & 0.303 / \small 0.000 & 0.339 / \small 0.000 & 0.283 / \small 0.000 & 0.315 / \small 0.000 & \hilight{0.290 / \small 0.000} & 0.305 / \small 0.000 & \textbf{0.346} / \small 0.000 & 0.314 / \small 0.000 & 0.323 / \small 0.000 & 0.297 / \small 0.000 & \hilight{0.337 / \small 0.000} & \hilite{0.316 / \small 0.000}\\
GPT-5.4 & 0.285 / \small 0.000 & 0.286 / \small 0.000 & \textbf{0.341} / \small 0.000 & 0.288 / \small 0.000 & 0.305 / \small 0.000 & \hilight{0.300 / \small 0.000 }& 0.324 / \small 0.000 & 0.335 / \small 0.000 & 0.317 / \small 0.000 & 0.303 / \small 0.000 & 0.296 / \small 0.000 &  \hilight{0.280 / \small 0.000} & \hilite{0.333 / \small 0.000}\\

\bottomrule
\end{tabular}}
\end{table}

\section{Case Study: How Table Representation Affects Attention} \label{sub: table-representation-attention}
\begin{figure}[H]
    \centering
    \includegraphics[width=0.9\linewidth]{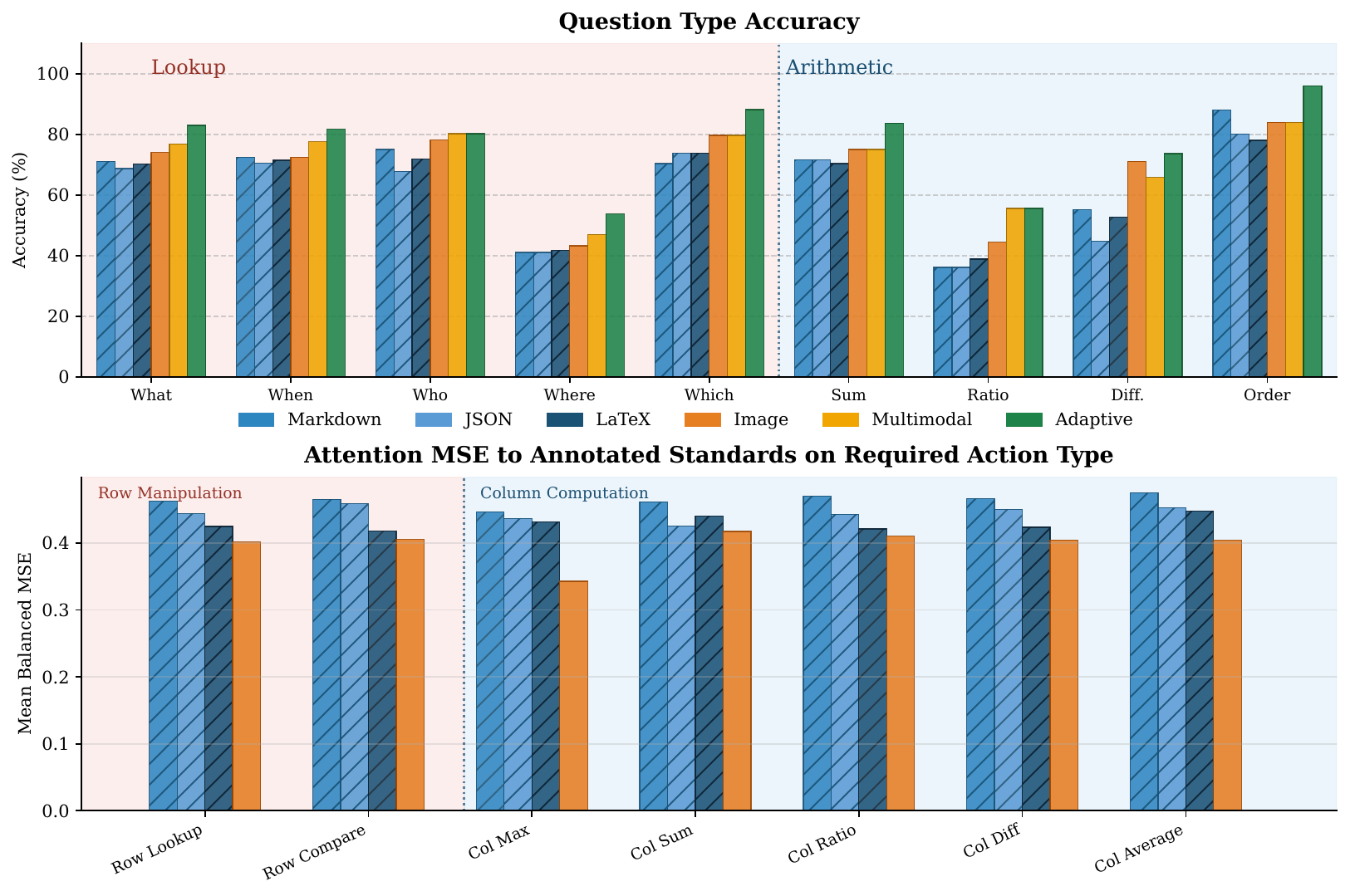}
    \caption{We annotate human-preferred attention standard, and visual-included representations not only lead to higher accuracy in different question types, but also lower MSE to human-preferred attention onto the table. This suggests that the relationship between more precise grounding over relevant table regions and quality of table reasoning.}
    \label{fig:placeholder}
\end{figure}
Figure~\ref{fig:main-ablation} reports a case study on how MLLM attends to the table when given queries requiring different actions. By extending the `Algeria' question from Figure~\ref{fig:atten-illus} into 20 \emph{1-step} questions, we annotate a human-preferred attention standard by labeling each cell in binary class to indicate relevancy to query. This allows us to measure the MSE between actual cell-level attention and human-preferred standard. Image modality consistently achieves lower MSE than text modalities, indicating more precise grounding over relevant table regions. Details of the questions are in Table~\ref{tab:tableqa-questions}. 

\begin{table*}[t]
\centering
\small
\setlength{\tabcolsep}{4pt}
\renewcommand{\arraystretch}{1.12}
\caption{Designed questions and required reasoning actions.}
\label{tab:tableqa-questions}
\begin{tabularx}{\textwidth}{X l X}
\toprule
\textbf{Question} & \textbf{Ground Truth} & \textbf{Required Actions} \\
\midrule
How many plants are in Algeria? & 6 & \texttt{sum\_over\_column} \\
What is the average capacity of Algeria plants? & 4.933 & \texttt{average\_over\_column} \\
What is the name and capacity of the plant in Brunei? & Lumut 1 with 7.2 mmtpa & \texttt{lookup\_row} \\
What is the average capacity of Malaysia plants? & 6.267 & \texttt{lookup\_row}, \texttt{average\_over\_column} \\
What is the latest plant built in Qatar? & 2010 & \texttt{lookup\_row}, \texttt{max\_over\_column} \\
What is the average capacity of all plants? & 8.508 & \texttt{average\_over\_column} \\
Which country has the most plants? & Algeria and Indonesia & \texttt{sum\_over\_column} \\
Are all plants in Egypt located in the same location? & No & \texttt{lookup\_row}, \texttt{comparison\_across\_rows} \\
Is the average capacity of Malaysian plants higher than that of Egyptian plants? & No & \texttt{lookup\_row}, \texttt{average\_over\_column}, \texttt{comparison\_across\_rows} \\
Which country has the highest-capacity plant? & Qatar & \texttt{max\_over\_column}, \texttt{lookup\_row} \\
What is the age difference between Algeria's oldest and newest plant? & 35 & \texttt{lookup\_row}, \texttt{difference\_within\_column} \\
Between Algeria's oldest and newest plants, how many times as large is the capacity of the newest plant as the oldest? & 6.667 & \texttt{ratio\_within\_column}, \texttt{lookup\_row} \\
Between plants from Angola and Brunei, which country's plant(s) has its location stated? & Angola & \texttt{comparison\_across\_rows} \\
Between the plant in Indonesia and Yemen, how many times as large is the capacity of the Indonesia plant as the Yemen plant? & 1.134 & \texttt{lookup\_row}, \texttt{ratio\_within\_column} \\
Which country has the newest plant? & Angola & \texttt{lookup\_row}, \texttt{comparison\_across\_rows} \\
Considering only plants with given capacity, is any Algeria plant larger in capacity than any Indonesia plant? & No & \texttt{lookup\_row}, \texttt{comparison\_across\_rows}, \texttt{difference\_within\_column} \\
What is the age difference between the oldest and newest plants in the table? & 49 & \texttt{difference\_within\_column} \\
What is the absolute capacity difference between the Brunei plant and the Nigeria plant? & 16.3 & \texttt{lookup\_row}, \texttt{difference\_within\_column} \\
What is the relative capacity difference between the Brunei plant and the Nigeria plant? & 2.264 & \texttt{lookup\_row}, \texttt{ratio\_within\_column} \\
Which country has the widest age gap between oldest and newest plant? & Algeria & \texttt{difference\_within\_column}, \texttt{lookup\_row}, \texttt{comparison\_across\_rows} \\
\bottomrule
\end{tabularx}
\end{table*}

\section{Case study on how different text serialization formats affect attention on tables}\label{sec:text-serialization-case-study}
Both empirical results in Table~\ref{tab:multimodal_perception} and Figure~\ref{fig:main-ablation}(a) indicate that LLM's understanding on table structure is sensitive to different text serialization formats, motivating us to conduct an in-depth attention-level case study. 

\begin{table}[H]
\centering
\caption{Canonical serialization templates for the same 3-column row.}
\label{tab:text-serialization-templates}
\begin{tabular}{lp{0.68\linewidth}}
\toprule
Format & Example serialization \\
\midrule
\texttt{JSON} & \texttt{[{"Col1":"val1","Col2":"val2","Col3":"val3"}]} \\
\texttt{LaTeX} & \texttt{begin\{tabular\}\{lll\} Col1 \& Col2 \& Col3 <ROW> val1 \& val2 \& val3 end\{tabular\}} \\
\texttt{Markdown} & \texttt{| Col1 | Col2 | Col3 | <ROW> | val1 | val2 | val3 |} \\
\bottomrule
\end{tabular}
\end{table}

We evaluate the WTQ~\citep{pasupat-liang-2015-compositional} using 4-bit quantized Qwen3-VL-8B-Instruct~\citep{bai2025qwen3vltechnicalreport}. For each table-question pair, we extract decoder attention from the answer-generation position and aggregate attention across all 36 layers of the chosen MLLM to measure the mean attention distribution over the prompt tokens. This allows us to examine how the MLLM agent distributes the attention budget across the table content as well as the format-specific structural tokens. We standardize the attention evaluation by measuring the attention ratio of different components of a TableQA query. Specifically, we have the following four evaluation diagnostics: \begin{itemize}
    \item Format overhead ratio that measures the fraction of table token budget spent on syntax $\frac{N_{\text{structural tokens}}}{N_{\text{table tokens}}}$
    \item Fraction of header attention to data attention to evaluate if the data rows receive sufficient attention: $\frac{\frac{1}{N_{\text{header tokens}}}\sum_{i\in\text{header tokens}}\mathrm{attn}(i)}{\frac{1}{N_{\text{data tokens}}}\sum_{j\in\text{data tokens}}\mathrm{attn}(j)}$
    \item Fraction of table attention to all tokens to evalaute how much prompt attention is directed to the table: $\frac{\frac{1}{N_{\text{table tokens}}}\sum_{i\in\text{table tokens}}\mathrm{attn}(i)}{\frac{1}{N_{\text{tokens}}}\sum_{j\in\text{tokens}}\mathrm{attn}(j)}$
    \item Table attention entropy which evaluates the Shannon entropy of attention on each content token to evaluate whether agent's attention is focused on specific cells, where a lower entropy means that the attention is more focused on specific cells
\end{itemize}

\begin{table}[ht]
\centering
\caption{LLM Accuracy (\%) by Token-Size Bin: Aggregated and Per-Dataset Breakdown}
\label{table: text-case-study}
\small
\setlength{\tabcolsep}{5pt}
\renewcommand{\arraystretch}{1.15}
\begin{tabular}{@{} l *{3}{c} @{}}
\toprule
\multicolumn{4}{c}{Overall Metrics (Correct Q / Wrong Q)}\\
\midrule
\textbf{Bin (Tokens)} & \textbf{JSON} & \textbf{LaTeX} & \textbf{Markdown} \\
\midrule
Accuracy (\%) & 72.2 & 70.7 & 71.2\\
Average Structural Token & 231 & 147 & 121 \\
Total Token & 469 & 416 & 417 \\
\midrule
Format overhead ratio (\%)
& \cellacc{51.8}{51/55}
& \cellacc{37.0}{38/34}
& \cellacc{46.9}{48/45}\\
Header-data attention ratio 
& \cellacc{1.29}{1.3/1.2}
& \cellacc{1.95}{2.0/1.7}
& \cellacc{2.10}{2.3/1.6}\\
Table attention fraction (\%) & \cellacc{51.6}{51/52} & \cellacc{36.3}{34/43} & \cellacc{36.7}{34/43} \\ 
Table attention entropy & \cellacc{4.71}{4.6/4.9} & \cellacc{5.13}{5.1/5.2} & \cellacc{5.14}{5.1/5.1} \\ \bottomrule
\bottomrule
\end{tabular}
\end{table}

We observe that the metric values between \texttt{LaTeX} and \texttt{Markdown} are more similar compared to \texttt{JSON}, e.g. the two formats have very similar table attention fraction, entropy and a more closely aligned header-data attention ratio, while format overhead ratio (header-data attention ratio) of \texttt{JSON} is higher (lower) than the other two formats. This result is not surprising as \texttt{JSON} has a distinctive format where it includes the column name for each serialized row (Table~\ref{tab:text-serialization-templates}, while the other two formats include the headers in the beginning and only record cell value in later parts, each separated by distinctive delimiters. The more closely aligned accuracy suggests that the agents are not sensitive to different delimiters, aligning with the results of previous works~\citep{kwok2026enhancingtableqaverifiablereasoning}. The performance deviation lies on the difference in formatting between \texttt{JSON} and the two other formats. Specifically, this repeated mentioning of column name in each row allows the MLLM to attend and understand each row separately, without the need of cross-referencing the column headers that are mentioned in a distant beginning. Despite resulting in more tokens needed to serialize the table, this aligns with the slightly outperformance in TableQA accuracy of \texttt{JSON} relative to \texttt{LaTeX} and \texttt{Markdown}. 

Our additional error localization analysis supports the above attribution, as we observe that \texttt{JSON} serialization has a 51\% and 55\% format overhead ratio for the correctly and incorrectly attempted questions respectively, where \texttt{LaTeX} and \texttt{Markdown} show a (38/34) and (48/45) percent respectively. This suggests that as table lengthens, the single mention of column header in \texttt{LaTeX} and \texttt{Markdown} reduces semantic information accessibility within rows, hence affecting the answer quality. Moreover, the remaining three ratios also indicate the repeated mention of header in each row contributes to better tabular understanding, whereas a lower header-data attention ratio indicates that the agent can connect each data entry with each header more closely, which then contributes to a better table attention fraction relative to non-table context in the prompt. The lower table attention entropy for \texttt{JSON} also suggests that having repeated mentioning of column headers improve semantic understandability of each row separately, hence allowing the agent to focus its attention better instead of attempting to revisit previous context to understand the cell values. These results suggest that by repeating the column header in each row, this facilitates MLLM understanding on tabular structure by preserving the semantic meaning of each row independently without the need to cross-attend distant headers. We also seek to understand whether these specific factors of different formatting would interact with different question types. 

\begin{figure}
    \centering
    \includegraphics[width=0.95\linewidth]{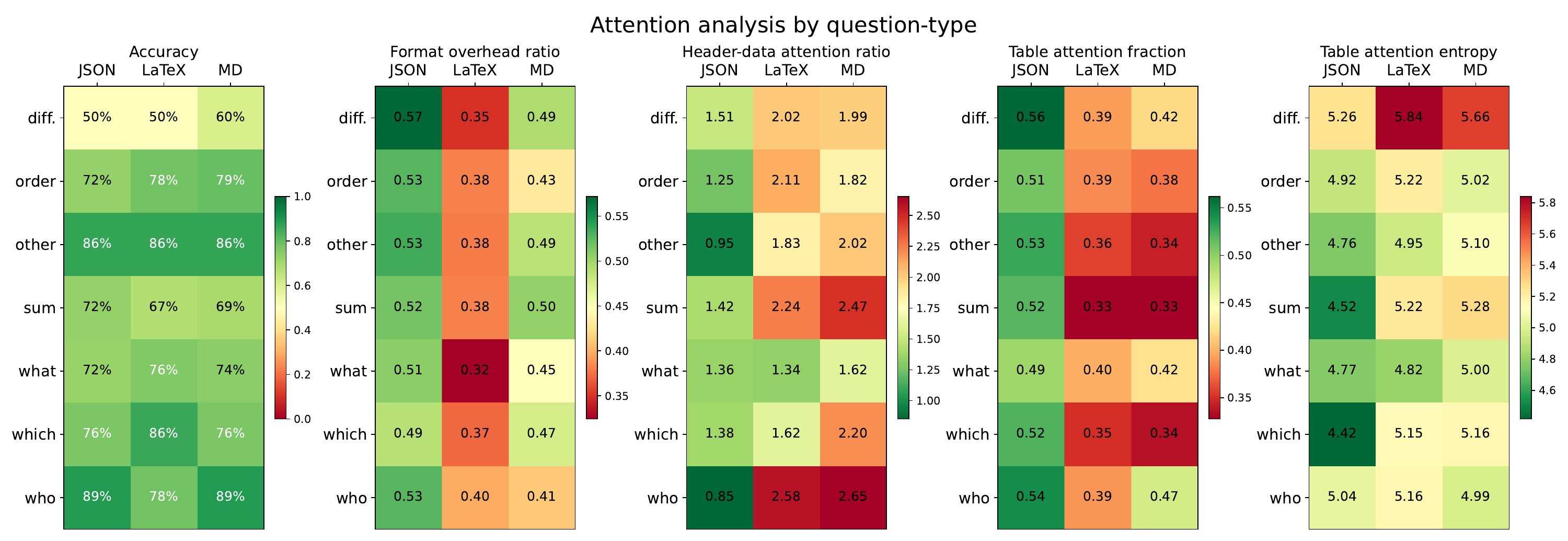}
    \caption{Heatmap of question-type performance and attention statistics across \texttt{JSON}, \texttt{LaTeX} and \texttt{Markdown} table serializations. Results show that while \texttt{JSON} devotes more attention to tables to improve lookup tasks, its heavier format overhead leads to underperformance in ordered computation tasks across multiple rows.}
    \label{fig:case_study_question_type}
\end{figure}

\paragraph{Break down performance by question types}
The question-type analysis on shows the weakness of the header-repeating \texttt{JSON}, as this action increases the contextual length of each row. While this facilitates unordered retrieval (illustrated by its outperformance in \texttt{sum} and \texttt{who}), it underperforms in tasks which require retrieving multiple row values (\texttt{which} and \texttt{order}). This suggests that as an increased structural syntax preserves individual semantic meaning of each row independently, it also results in more attention being absorbed by structural syntax instead of semantic content, hence hurting performance that require multi-row retrievals where the JSON fails to disperse the attention effectively across more cell values. The result further suggests potential values of leveraging different formats for table serialization to further diversify action-conditioned multimodal selection policy, motivating further explorations on integrating different modalities and formats to improve MLLM agents' tabular understanding ability. 

\section{Case Study: How Table Size Affects Modality Performance}
\begin{table}[ht]
\centering
\caption{LLM Accuracy (\%) by Token-Size Bin: Aggregated and Per-Dataset Breakdown}
\label{table: accuracy-by-table}
\small
\setlength{\tabcolsep}{5pt}
\renewcommand{\arraystretch}{1.15}
\begin{tabular}{@{} l *{6}{c} @{}}
\toprule
\multicolumn{7}{c}{Overall Accuracy (\%) (WTQ/MMQA/MMTU)}\\
\midrule
\textbf{Bin (Tokens)} & \textbf{JSON} & \textbf{LaTeX} & \textbf{Markdown} & \textbf{Image} & \textbf{Multimodal} & \textbf{Adaptive} \\
\midrule
41--263
& \cellacc{59.5\%}{80/60/21}
& \cellacc{60.2\%}{82/59/21}
& \cellacc{58.3\%}{81/59/20}
& \cellacc{62.8\%}{85/63/30}
& \cellacc{63.1\%}{84/63/30}
& \cellacc{\textbf{70.1\%}}{91/68/41} \\
264--375
& \cellacc{74.5\%}{83/65/75}
& \cellacc{75.0\%}{85/67/79}
& \cellacc{76.9\%}{85/70/75}
& \cellacc{77.7\%}{86/63/82}
& \cellacc{80.0\%}{85/72/82}
& \cellacc{\textbf{83.5\%}}{88/74/86} \\
376--510
& \cellacc{73.5\%}{88/67/74}
& \cellacc{74.8\%}{84/66/74}
& \cellacc{76.1\%}{89/68/86}
& \cellacc{74.3\%}{88/69/81}
& \cellacc{74.8\%}{88/67/86}
& \cellacc{\textbf{78.5\%}}{92/73/83} \\ 
513--816
& \cellacc{64.9\%}{69/62/30}
& \cellacc{66.8\%}{70/65/33}
& \cellacc{69.7\%}{73/69/34}
& \cellacc{69.0\%}{73/64/34}
& \cellacc{70.7\%}{74/65/38}
& \cellacc{\textbf{75.5\%}}{79/69/43} \\
817--1695
& \cellacc{67.3\%}{74/67/27}
& \cellacc{66.0\%}{73/66/31}
& \cellacc{68.9\%}{75/66/32}
& \cellacc{71.4\%}{80/68/34}
& \cellacc{72.9\%}{78/69/38}
& \cellacc{\textbf{77.8\%}}{82/77/42} \\ 
1720--94774
& \cellacc{45.7\%}{73/47/31}
& \cellacc{45.4\%}{72/48/24}
& \cellacc{47.0\%}{71/48/32}
& \cellacc{48.7\%}{72/50/37}
& \cellacc{51.0\%}{73/56/37}
& \cellacc{\textbf{55.7\%}}{77/62/39} \\ 
\bottomrule
\bottomrule
\end{tabular}
\end{table}

To understand when different table representations succeed or fail, we conduct a case study on the effect of table size on reasoning accuracy. We measure table size by the number of serialised tokens, as some tables contain long text cells, where trivial dimensionality fail to quantify contextual length within a cell. Within our evaluated datasets and models, we observe a non-monotonic relatoinship between table size and accuracy in Table~\ref{table: accuracy-by-table}. Specifically, we observe relatively low performance in smallest token bin. The performance peaks at second smallest token bin and declines steadily as the table becomes larger. Specifically, we observe that aside from the proposed action-conditioned multimodal selection policy, which remains outperforming all other fixed modalities as shown in Figure~\ref{fig:main-ablation}, visual-grounded modalities such as image and multimodal show more graceful performance degradation. We attribute this to visual rendering which partially compresses the table before reasoning, releasing some of the attention burden on the symbolic delimiters for table structuring, where are unavoidable for text serialization formats. Regarding the non-trivial underperformance in the smallest bin in contrary to the decreasing trend, we conduct a more in-depth study on each dataset. 

\paragraph{Break down performance by datasets}
By analyzing the performance on each dataset separately, we observe that this low accuracy in the smallest table size originates from a table composition over the three datasets. Specifically, in that bin, the MMTU with least accuracy contributes with a relatively higher percentage of 31\%. In comparison, WTQ contributes to only 20\% of that bin while attaining a 91\% accuracy. More specifically, by analyzing the failure cases in MMTU, we observe that these small MMTU tables contain non-trivial TableQA scenarios, with compact tabular fragments requiring dense numerical reasoning, such as percentage growth over two rows while ambiguous keywords in question so that the agent is unable to retrieve the respective column directly from the header. In the contrary, WTQ and MMQA tables contain more standardized tables, facilitating table understanding of the agent.

\end{document}